\documentclass[times, preprint,10pt]{elsarticle}

\usepackage{amssymb}
\usepackage{booktabs}
\usepackage{subcaption}
\usepackage{url}
\usepackage{xcolor}
\usepackage{comment}
\usepackage{multirow}

\usepackage{amsmath}

\begin{document}
\emergencystretch 3em

\begin{frontmatter}



\title{ProtoX-AD: Self-Explainable Time Series Anomaly Detection and Characterization}


\author[1]{Aitor Sánchez-Ferrera\corref{cor1}}
\ead{aitor.sanchezf@ehu.eus}

\author[2]{Elisabeth Wetzer \fnref{fn1}}

\author[2]{Kristoffer Wickstrøm\fnref{fn1}}

\author[2]{Michael Kampffmeyer\fnref{fn1}}

\author[2]{Robert Jenssen\fnref{fn1}}

\affiliation[1]{
    organization={Department of Computer Science and Artificial Intelligence, University of the Basque Country UPV/EHU},
    city={Donostia},
    country={Spain}
}

\affiliation[2]{organization={Department of Physics and Technology, UiT The Arctic University of Norway},
    city={Tromsø},
    country={Norway}}

\cortext[cor1]{Corresponding author.}
\fntext[fn1]{These authors contributed equally to this work.}

\begin{abstract}
Recent advances in time series anomaly detection (TSAD) have highlighted the effectiveness of self-supervised classification-based approaches. These methods apply transformations to normal training samples, training a classifier to recognize transformation-specific patterns that help identify anomalies through increased classification errors. Despite their strong performance, a significant challenge is their lack of explainability, as they provide limited insight into the characteristics of flagged anomalies. To address this limitation, we propose ProtoX-AD, a prototype-based self-explainable framework for self-supervised TSAD. ProtoX-AD learns transformation-aware latent representations alongside interpretable prototypes, enabling both accurate anomaly detection and the identification of distinct anomalous profiles through prototype-based explanations. Additionally, it allows for systematic analysis of how transformation design impacts detection performance and explainability. Experimental results on synthetic and real-world datasets demonstrate that ProtoX-AD achieves detection performance comparable to its black-box counterparts while offering more consistent and semantically meaningful explanations than existing explainable baselines. Our code is publicly
available at \url{https://github.com/Aitorzan3/ProtoX-AD}
\end{abstract}

\begin{keyword}
Time series anomaly detection \sep anomaly characterization \sep self-supervised learning \sep explainable AI \sep prototype learning.

\end{keyword}

\end{frontmatter}



\section{Introduction}\label{sec:introduction}

Time series anomalies are abnormal events in which the behavior of a system departs from its regular operating patterns~\citep{carreno2020analyzing}. Time series anomaly detection (TSAD) is a fundamental problem in many critical application domains, such as finance~\citep{hilal2022financial}, Internet of Things (IoT) systems~\citep{cook2019anomaly}, and healthcare~\citep{yang2023deep}. Due to the high cost of labeling anomalies, TSAD is typically addressed using unsupervised approaches~\citep{blazquez2021review,zamanzadeh2024deep}. Within this setting, models learn the patterns of normal data to detect deviations from normality at inference time, typically quantified as anomaly scores~\citep{choi2021deep}.

Recent advances in the field have focused on self-supervised learning~\citep{liu2021self}, with classification-based approaches emerging as a prominent paradigm~\citep{sanchez2025review}. In this setting, a set of transformations is applied to normal samples to generate one augmented view per transformation, and a classifier is trained to predict the transformation associated with each view~\citep{hojjati2022self}. As a result, models capture the expected response of normal samples when subjected to each transformation. At inference time, anomalies tend to violate these learned patterns, leading to higher classification errors that are used as anomaly scores. The effectiveness of self-supervised classification-based time series anomaly detection (SSC-TSAD) critically depends on the choice of transformations~\citep{yoo2022data}. 

Despite their strong performance, these approaches lack explainability~\citep{lee2024explainable}, limiting the understanding of both how transformation design influences detection performance and how different anomalous patterns are characterized. In particular, existing methods typically provide anomaly scores without offering insight into the underlying structure of anomalous behavior. Consequently, they do not support the identification and interpretation of distinct anomalous profiles, which is critical in many real-world applications~\citep{foorthuis2021nature}.

To address this limitation, in this paper we propose a self-explainable framework for SSC-TSAD\footnote{In this work, we focus on detecting entire anomalous time series as opposed to detecting anomalous points or subsequences.}. We draw inspiration from prototypical self-explainable models (SEMs) in the field of eXplainable AI (XAI)~\citep{bai2021explainable}, which learn representative concepts in the latent space that can be visualized in the input space to explain model decisions~\citep{chen2019looks, gautam2023looks}. By structuring the latent space around transformation-induced concepts and reconstructing prototypes in the input domain, our method enables both accurate anomaly detection and the characterization of distinct anomalous profiles. Moreover, it enables a systematic analysis of the impact of transformation design on SSC-TSAD. We call our method ProtoX-AD.

In summary, the main contributions of this work are as follows:

\begin{itemize}
    \item We propose ProtoX-AD, a prototype-based self-explainable framework for self-supervised classification-based time series anomaly detection.
    
    \item Our method achieves anomaly detection performance comparable to its black-box self-supervised counterpart, while providing interpretable prototype-based explanations that enable anomaly characterization.
    
    \item We provide an analysis of the impact of transformation design on SSC-TSAD in terms of detection performance, explainability, and anomaly characterization.
\end{itemize}

\section{Related Work on SSC-TSAD}

Self-supervised classification-based TSAD methods rely on the application of a set of transformations to normal samples, generating augmented views that define a surrogate classification task~\citep{gui2024survey}. The effectiveness of these approaches depends strongly on the choice of transformations, as they determine the structure of the learned representations and the types of deviations that can be detected. Specifically, recent work states that they should generate corrupted views of normal samples that reflect the properties of the targeted anomalies~\citep{yoo2022data}.

Following this perspective, most existing approaches rely on manually designed transformations tailored to specific anomaly detection problems based on domain knowledge\cite{yoo2022data}. For instance, transformations such as upscaling have been used in water leak detection to mimic increased consumption patterns caused by leaks~\citep{blazquez2021water}. Similarly, various works propose transformations that alter the amplitude and frequency of sequences to mirror the properties of seizures for epilepsy detection~\citep{zheng2022task, xu2020anomaly}. Although effective, relying on hand-crafted transformations limits the general applicability and transferability of these methods across TSAD problems~\citep{sanchez2025review}.

To alleviate this limitation, recent works have explored learnable neural transformations~\citep{qiu2021neural, sanchez2025neucoreclass}, implemented as neural networks that are learned directly from normal training data. By leveraging contrastive learning~\citep{hu2024comprehensive}, these transformations are enforced to generate diverse and non-redundant augmented views while preserving the semantic content of the original samples and disrupting data normality in a controlled manner, prioritizing general applicability over domain-specific assumptions~\citep{sanchez2025neucoreclass}. 

The choice between manual and neural transformations often depends on the availability of domain knowledge: manually designed transformations can target specific anomalous patterns, whereas neural transformations aim to provide greater generality~\cite{sanchez2025review}. Existing work typically compares these approaches in terms of anomaly detection performance, highlighting the trade-off between domain specificity and generality. However, this perspective does not capture how different transformation choices influence the characterization of anomalous behavior. In particular, understanding how transformed views relate to distinct anomalous patterns requires insight beyond a single anomaly score. Existing methods lack explicit mechanisms to support such analysis~\cite{lee2024explainable}.

\section{Method}

ProtoX-AD learns a latent space in which augmented views are well suited for SSC-TSAD alongside prototypes that allow the distinction and characterization of different anomalous profiles through prototypical explainability.

\subsection{Model Architecture}\label{sec:architecture}

\begin{figure*}[t]
    \centering
    \includegraphics[width=0.9\textwidth]{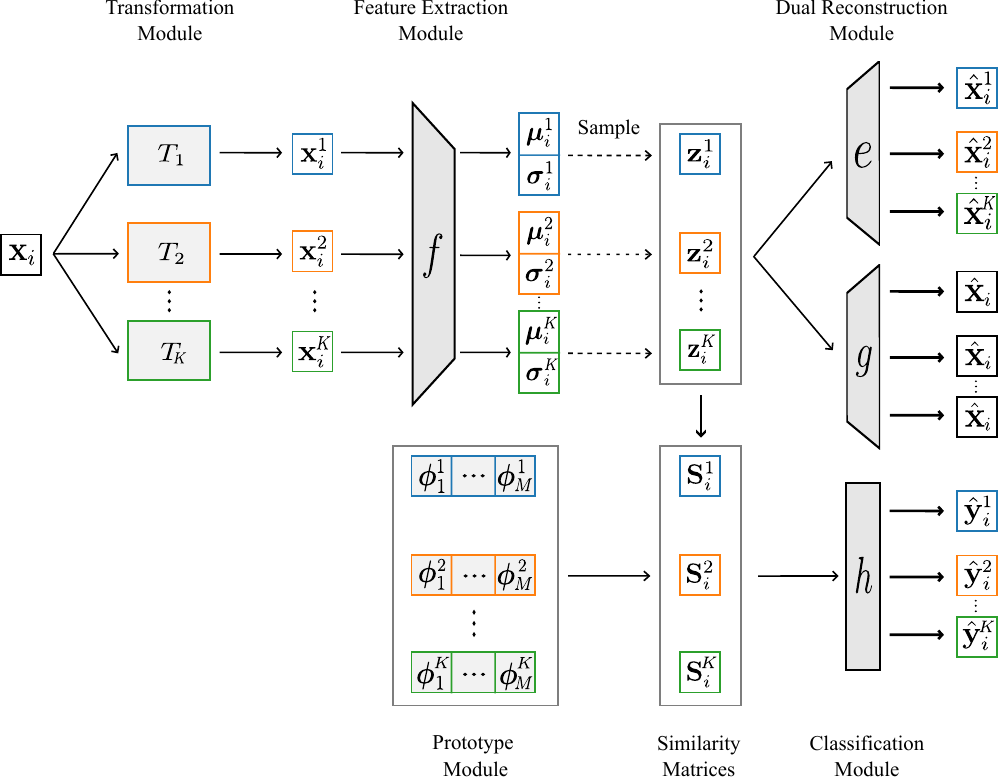}
    \caption{Pipeline of ProtoX-AD. The input time series $\mathbf{x}_i$ is transformed into augmented views $\mathbf{x}_i^k$, encoded into latent representations $\mathbf{z}_i^k$, and compared with class-specific prototypes $\boldsymbol{\phi}_m^k$ to compute similarity matrices $\mathbf{S}_i^k$. The classification module performs self-supervised classification based on these similarities. In addition, the dual reconstruction module reconstructs both transformed views and the original sample from the latent representations.}
    \label{fig:pipeline}
\end{figure*}

Our model’s pipeline consists of five key components: (i) a transformation module, (ii) a feature extraction module, (iii) a dual reconstruction module, (iv) a prototype module, and (v) a classification module. Figure~\ref{fig:pipeline} illustrates the overall architecture of ProtoX-AD.

\paragraph{Transformation module} This comprises a set of $K$ transformations $\mathcal{T} = \lbrace T_1, ..., T_K \rbrace$, which can be manual or learnable neural transformations representing parameterized functions. When applied to an input sample $\mathbf{x}_i$, these transformations generate a set of augmented views associated with the original sample $\mathcal{T}(\mathbf{x}_i) = \lbrace T_1(\mathbf{x}_i), ..., T_K(\mathbf{x}_i) \rbrace = \lbrace \mathbf{x}_i^1, ..., \mathbf{x}_i^K \rbrace$, where $\mathbf{x}_i^k$ denotes the augmented view that is generated by applying the $k$-th transformation $T_k(\cdot)$ to that sample. Following the existing literature on self-supervised classification for anomaly detection~\cite{sanchez2025review}, the first transformation $T_1(\cdot)$ in the transformation module is set to the identity function, i.e., $T_1(\mathbf{x}_i) = \mathbf{x}_i$.

\paragraph{Feature extraction module} This consists of an encoder $ f(\cdot) $ that maps the augmented views into a new representation that is suitable for the self-supervised anomaly detection task. Following the variational autoencoder paradigm and inspired by ProtoVAE~\cite{gautam2022protovae}, ProtoX-AD models the latent space probabilistically to encourage smooth and structured latent representations. Specifically, each augmented view $\mathbf{x}_i^k$ is transformed by $f(\cdot)$ into a tuple  ($\boldsymbol{\mu}_i^k$, $\boldsymbol{\sigma}_i^k$), which are the parameters of the posterior normal distribution $\mathcal{N}$($\boldsymbol{\mu}_i^k$, $\boldsymbol{\sigma}_i^k$), from where its corresponding latent representation $\mathbf{z}_i^k$ is sampled. 

\paragraph{Dual reconstruction module}
It consists of an explainer $e(\cdot)$ and a semantic preservation decoder $g(\cdot)$, which are designed to be symmetric to the encoder $f(\cdot)$. The explainer $e(\cdot)$ reconstructs the augmented views of each sample $\mathbf{x}_i^k$ from their corresponding latent representations $\mathbf{z}_i^k$, while the semantic preservation decoder $g(\cdot)$ reconstructs the original sample $\mathbf{x}_i$ from each latent view $\mathbf{z}_i^k$. That is, ideally $e(\mathbf{z}_i^k) = \mathbf{x}_i^k$ and $g(\mathbf{z}_i^k) = \mathbf{x}_i$, for $k \in \{1, 2, \dots, K\}$.

At a high level, the explainer $e(\cdot)$ enables the visualization and interpretation of latent prototypes in the input space, while the semantic preservation decoder $g(\cdot)$ acts as a regularizer that preserves semantic consistency across transformations and prevents representation collapse.

\paragraph{Prototype module} Our network incorporates a set of prototypes that are learned to capture representative concepts associated with each class and to enable the self-supervised classification task. Specifically, we consider $M$ prototypes for each class, resulting in a set of prototypes $\boldsymbol{\Phi} = \lbrace \boldsymbol{\phi}_{m}^k\rbrace_{m=1..M} ^{k=1..K}$, where $\boldsymbol{\phi}_{m}^k$ is the $m$-th prototype associated with class $k$.  For each augmented view in the latent space $\mathbf{z}_i^k$, we define a matrix of distances $\mathbf{D}_i^k \in \mathbb{R}^{K \times M}$, whose entry $d_i^k(c,m)$ corresponds to the squared Euclidean distance between $\mathbf{z}_i^k$ and the $m$-th prototype of class $c$:

\begin{equation}
    d_i^k(c,m) = \lVert \mathbf{z}_i^k - \boldsymbol{\phi}_m^c \rVert_2^2.
\end{equation}

The prototype vectors are initialized using K-means clustering in the latent space. Specifically, prior to training, an initial forward pass of the training data through the transformation module and encoder is performed. Then, K-means with $M$ clusters is applied independently per transformation class to the resulting latent representations to obtain an initial set of latent prototypes.

\paragraph{Classification module}

To ensure transparency, the self-supervised classification task is performed by directly leveraging the concepts learned in the model’s latent space. For each augmented view with latent representation $\mathbf{z}_i^k$, we convert its associated distance matrix $\mathbf{D}_i^k$ into a similarity matrix $\mathbf{S}_i^k$ by taking its negative, i.e., $\mathbf{S}_i^k = -\mathbf{D}_i^k$. Accordingly, the entries of the similarity matrix are defined as

\begin{equation}
    s_i^k(c,m) = - d_i^k(c,m),
\end{equation}

\noindent
where $s_i^k(c,m)$ quantifies the similarity between the latent representation $\mathbf{z}_i^k$ and the $m$-th prototype of class $c$.

Following~\cite{chen2019looks}, the resulting similarities are then fed into a linear classifier $h(\cdot)$ to compute the self-supervised classification predictions for each augmented view, yielding an output $\hat{\mathbf{y}}_i^k = h(\mathbf{S}_i^k)$ that ideally assigns the corresponding transformation label $k$, for $k \in \{1, 2, \dots, K\}$.

\subsection{Training ProtoX-AD} \label{sec:training}

ProtoX-AD is trained using a dataset of unlabeled time series samples $\mathcal{X} = \{\mathbf{x}_i\}_{i=1}^N$, where (ideally) all the samples belong to the class designated as normal. The overall loss function of the model is defined as:

\begin{equation}
\mathcal{L}_{\textnormal{ProtoX-AD}} = \mathcal{L}_{\textnormal{class}} + \mathcal{L}_{\textnormal{recon}} + 
\mathcal{L}_{\textnormal{proto}}.
\end{equation}

We now detail each term.

\paragraph{Classification through prototypes}
This learning objective enables the model to distinguish the transformations considered in the transformation module, thereby learning the self-supervised classification task. We employ the standard cross-entropy loss defined as

\begin{equation}
    \mathcal{L}_{\textnormal{class}} = - \dfrac{1}{NK} \sum_{i=1}^N \sum_{k=1}^K \mathbf{y}_i^k \log(\hat{\mathbf{y}}_i^k),
\end{equation}

\noindent
where $\mathbf{y}_i^k \in \{0,1\}^K$ is the corresponding self-supervised one-hot label
with $y_i^k(k)=1$ and $0$ otherwise.

The latent representations associated with different transformations are separated in the latent space, promoting transformation diversity. Since the transformation associated with the first class is defined as the identity function, this objective further promotes the generation of augmented views that disrupt data normality, as transformed views are encouraged to deviate from the latent structure of normal samples.

\paragraph{Dual reconstruction training objective}

The dual reconstruction training objective is designed to support SSC-TSAD by jointly structuring the latent space around transformation-specific concepts and preserving the semantic identity of the original samples. This learning objective is composed of three complementary terms.

The first term enables the explainer $e(\cdot)$ to reconstruct each augmented view from its latent representation by minimizing:

\begin{equation}
    \mathcal{L}_{\textnormal{exp}} =
    \frac{1}{N K} \sum_{i=1}^N \sum_{k=1}^K 
    \left\lVert e(\mathbf{z}_i^k) - \mathbf{x}_i^k \right\rVert_1.
\end{equation}

The second term enforces the latent representations of augmented views to preserve the semantic information of their associated original samples through the semantic preservation decoder $g(\cdot)$:

\begin{equation}
    \mathcal{L}_{\textnormal{sem}} =
    \frac{1}{N K} \sum_{i=1}^N \sum_{k=1}^K
    \left\lVert g(\mathbf{z}_i^k) - \mathbf{x}_i \right\rVert_1.
\end{equation}

In addition to reconstruction-based objectives, we further structure the latent space around transformation-specific concepts by introducing a prototype-centered latent regularization inspired by the mixture-based formulation of~\cite{gautam2022protovae}:

\begin{equation}
\mathcal{L}_{\textnormal{mix}} = 
\frac{1}{N K}
\sum_{i=1}^N \sum_{k=1}^K
\sum_{m=1}^M
w_i^k(m)\,
D_{KL}\!\left(
\mathcal{N}(\boldsymbol{\mu}_i^k, \boldsymbol{\sigma}_i^k)
\,\|\, 
\mathcal{N}(\boldsymbol{\phi}_{m}^k, \mathbf{I}_d)
\right),
\end{equation}

\noindent
where $d$ is the dimensionality of the latent space, $\mathbf{I}_d$ denotes the $d \times d$ identity matrix, $D_{KL}$ is the Kullback--Leibler divergence, $\mathcal{N}$ denotes a multivariate normal distribution, and the mixture weights $w_i^k(m)$ are defined as

\[
w_i^k(m)=
\frac{\exp\!\left(s_i^k(k,m)\right)}
{\sum_{m'=1}^{M}\exp\!\left(s_i^k(k,m')\right)}.
\]

The overall training objective of the dual reconstruction module is given by
\begin{equation}
    \mathcal{L}_{\textnormal{recon}} =
    \mathcal{L}_{\textnormal{exp}} + \mathcal{L}_{\textnormal{sem}} + \mathcal{L}_{\textnormal{mix}}.
\end{equation}

This combined objective grounds the latent space in the input domain by jointly enforcing complementary constraints. The explainer and semantic reconstruction terms preserve input-level fidelity and latent intra-class diversity across augmented views, respectively. In turn, the mixture-based regularization aligns latent representations with class-specific prototypes, stabilizing prototype learning and discouraging degenerate solutions. Together, these components ensure that the learned latent space remains both structured and semantically interpretable.

\paragraph{Meaningful prototype learning}

To encourage meaningful and well-supported prototypes, we adopt two complementary objectives: a clustering loss and a prototype coverage loss.

Inspired by~\cite{chen2019looks}, the clustering loss encourages the latent representations of each augmented view to be close to at least one prototype related to its class, and it is defined as:

\begin{equation}
    \mathcal{L}_{\textnormal{clst}} =
    \frac{1}{NK}\sum_{i=1}^{N}\sum_{k=1}^{K}\min_{m\in\{1,\dots,M\}} d_i^k(k,m).
\end{equation}

Complementarily, the prototype coverage loss follows the intuition of~\cite{li2023prototypes} and ensures that each prototype is close to at least one augmented view associated with its class in the latent space. It is defined as:

\begin{equation}
    \mathcal{L}_{\textnormal{cov}} =
\frac{1}{KM}\sum_{k=1}^{K}\sum_{m=1}^{M}\min_{i\in\{1,\dots,N\}} d_i^k(k,m).
\end{equation}

The overall objective for meaningful prototype learning is then given by

\begin{equation}
\mathcal{L}_{\textnormal{proto}} =
\mathcal{L}_{\textnormal{clst}} +
\mathcal{L}_{\textnormal{cov}}.
\end{equation}

This combined objective enforces a meaningful prototypical structure in the latent space. The clustering loss promotes compactness of latent representations around class-specific prototypes, thereby aligning augmented views with representative concepts. In addition, the coverage loss prevents the emergence of unused or degenerate prototypes, a phenomenon often referred to as \textit{ghosting}~\cite{gautam2023prototypical}.

\subsection{Anomaly Detection with ProtoX-AD} \label{sec:detection}

After training, when evaluating a new sample $\mathbf{x}_{\textnormal{new}}$, ProtoX-AD encodes the identity augmented view of the sample into the latent space using the feature extraction module. The anomaly score of the new sample $\textnormal{AS}(\mathbf{x}_{\textnormal{new}})$ is defined as the cross-entropy loss associated with its self-supervised classification:

\begin{equation}
    \textnormal{AS}(\mathbf{x}_{\textnormal{new}})  = -   \mathbf{y}_{\textnormal{new}}^1 \log(\hat{\mathbf{y}}_{\textnormal{new}}^1),
\end{equation}

\noindent
where $\hat{\mathbf{y}}_{\textnormal{new}}^{1}$ denotes the model prediction for the identity view of $\mathbf{x}_{\textnormal{new}}$, and $\mathbf{y}_{\textnormal{new}}^{1}$ is the one-hot pseudo-label corresponding to the identity transformation (i.e., class $k=1$).

The underlying rationale is that the model is trained to associate each transformation with a specific class and to capture the corresponding transformation-induced patterns in the latent space. When evaluating a new sample, if its identity view is classified as a different transformation, this indicates that the sample exhibits characteristics that resemble those induced by that transformation. Consequently, a high cross-entropy value for the identity view reflects a semantic mismatch between the sample and the learned normal transformation-specific patterns, and therefore a higher likelihood that the sample is anomalous.

\subsection{Explanation through Prototypes} 

After training, the set of learned prototypes $\boldsymbol{\Phi}$ can be visualized in the input space by decoding them through the explainer network $e(\cdot)$. These decoded prototypes represent global concepts associated with the different classes defined by the transformations considered in the transformation module of ProtoX-AD.

To explain the anomaly score computation of a new sample $\mathbf{x}_{\textnormal{new}}$, we anchor the explanation to its identity view, which provides a canonical and human-interpretable representation of the sample. The explanation is obtained by identifying the nearest prototype to the identity view in the latent space and visualizing it in the input space as a representative concept capturing the normal or anomalous characteristics of the sample. In addition, the diversity induced by the transformation design enables the characterization of different anomalous profiles, as anomalous samples can be associated with prototypes corresponding to distinct transformation-specific concepts.

\subsection{ProtoX-AD is a SEM}

A model can be considered self-explainable if it fulfills three key properties: (i) \emph{transparency}, where concepts are directly used to perform the downstream task and are visualizable in the input space; (ii) \emph{diversity}, where concepts represent non-overlapping information in the latent space; and (iii) \emph{trustworthiness}, where performance matches that of a comparable black-box model and explanations are consistent, i.e., similar inputs yield similar explanations~\cite{gautam2022protovae}.

ProtoX-AD satisfies these properties by explicitly embedding interpretable concepts into its learning objective. It is transparent, as anomaly detection is performed through prototype-based classification, and prototypes are reconstructed in the input space using the explainer network. It promotes diversity by learning multiple prototypes per transformation-induced class that capture distinct variations of the underlying signal. Finally, ProtoX-AD is trustworthy: it achieves competitive performance with respect to its black-box counterpart (see Section~\ref{sec:experiments}), and the structured latent space, enforced by the VAE formulation, ensures that similar samples are mapped to similar prototypes, yielding consistent explanations.

\section{Experimental Setup}

In this section, we describe the experimental setup used to evaluate both the anomaly detection performance and the explainability capabilities of ProtoX-AD across different datasets.

\subsection{Datasets and problem definition}\label{sec:datasets}

We consider three TSAD problems based on synthetic and real-world time series data.

\begin{figure}[t]
    \centering
    \includegraphics[width=0.75\linewidth]{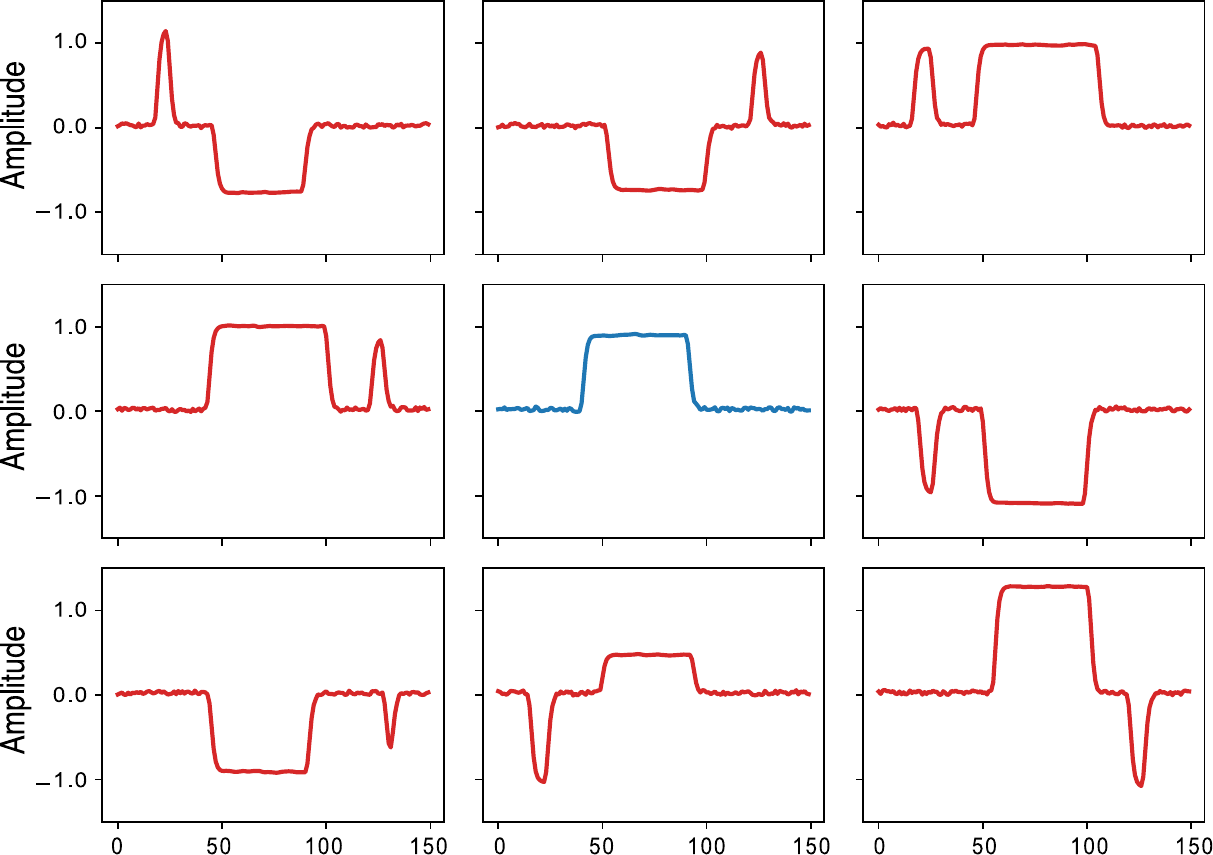}
    \caption{Representative normal (blue) and anomalous (red) time series from the UMD dataset.}
    \label{fig:samples_umd}
    \vspace{0.6em}
\end{figure}

\paragraph{UMD Dataset}
This dataset consists of synthetic time series sharing a common baseline structure with a central plateau. It comprises three classes: one class without bell-shaped patterns and two classes containing upward and downward bell-shaped events, respectively. Sequences with bell-shaped patterns are treated as anomalous samples, where these may appear either at the beginning or end, and the central plateau may appear inverted (see Figure~\ref{fig:samples_umd}). Predefined training and evaluation splits follow ~\cite{dau2019ucr}.

Following the unsupervised TSAD setting, the training set is restricted to normal samples only, while the evaluation set contains both normal and anomalous samples.

\paragraph{Global Temperature Anomalies Dataset (GTA)}
The dataset consists of monthly-sampled time series derived from global mean surface temperature anomaly data obtained from the GISTEMP~\cite{lenssen2024nasa} and gcag~\cite{noaa_gcag} sources. Each time series corresponds to a single year of the deviation of the monthly global temperature with respect to a reference climatological baseline, where positive values indicate warmer-than-average conditions and negative values indicate colder-than-average conditions. Each year is associated with an overall annual temperature anomaly, ranging approximately between $-1.5$ and $+1.5$ degrees.

\begin{figure}[t]
    \centering
    \includegraphics[width=\linewidth]{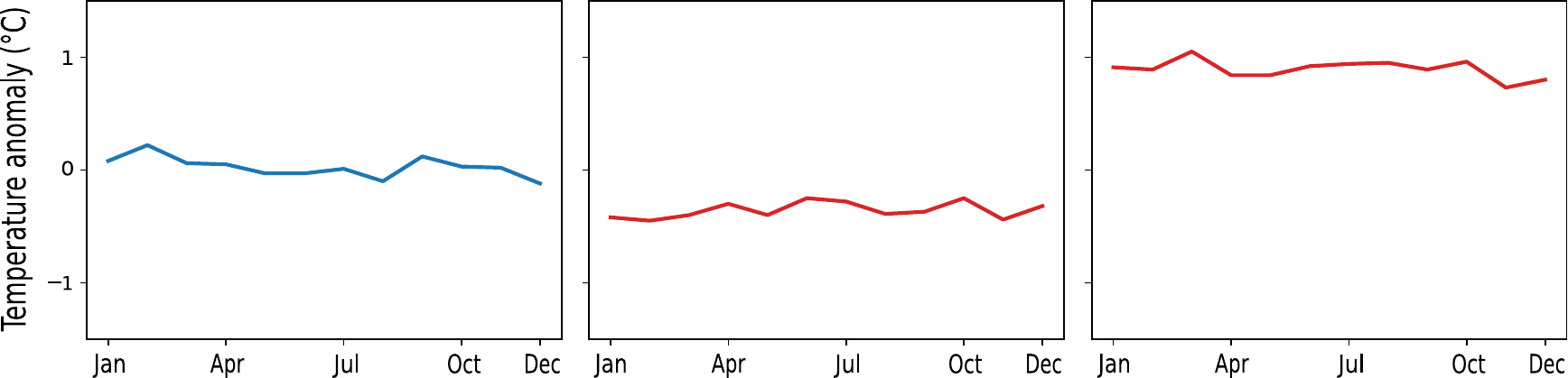}
    \caption{Representative normal (blue) and anomalous (red) yearly temperature anomaly time series from the GISTEMP source.}
    \label{fig:samples_gta}
\end{figure}

Years with an annual temperature anomaly within the range $[-0.25, 0.25]$ are treated as normal while remaining years are considered anomalous. We use $80\%$ of the normal samples for training, and the remaining $20\%$ of normal samples and anomalous samples for evaluation. In our experiments, the GISTEMP and gcag sources are treated as two independent TSAD problems and evaluated separately. Figure~\ref{fig:samples_gta} illustrates representative normal and anomalous yearly temperature anomaly time series from the GISTEMP source.

\paragraph{Yorkshire Water Leak Detection Dataset}
This is a real-world dataset containing water flow measurements across more than 2000 Distribution Management Areas (DMAs) in Yorkshire. We construct one time series per DMA and day, capturing the minimum night flow (MNF) behavior, and restrict our analysis to five DMAs that do not contain invalid or missing MNF values and exhibit consistent measurement quality, following the experimental protocol of~\cite{blazquez2021water}. We consider DMAs 549, 913, 1164, 1406, and 1259.

\begin{figure}[t]
    \centering
    \includegraphics[width=\linewidth]{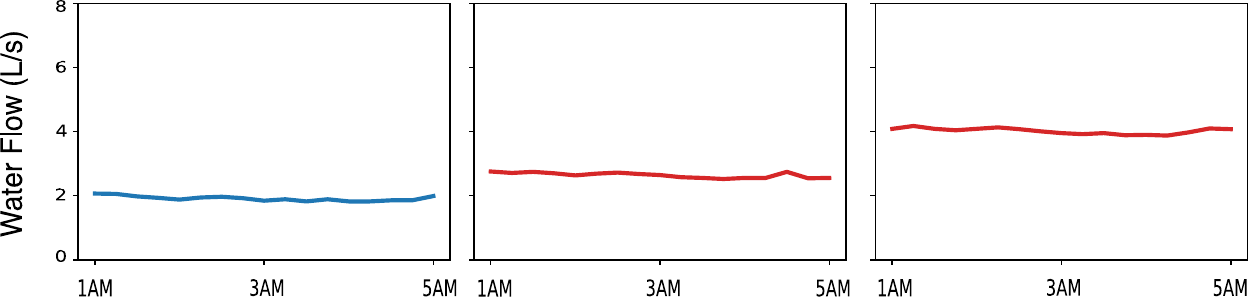}
    \caption{Representative normal (blue) and anomalous (red) water flow sequences from the Yorkshire Water Leak Detection dataset.}
    \label{fig:samples_yorkshire}
\end{figure}

Anomalies are defined based on percentile thresholds computed over the MNF values, following~\cite{blazquez2021water}. Percentile values of 80, 85, 90, and 95 are considered, leading to four experimental settings. Each combination of DMA, weekday, and percentile threshold defines an independent TSAD problem, resulting in a total of 140 TSAD problems. For each, $80\%$ of the normal samples are used for training, and the remaining $20\%$ of normal samples and all anomalous samples constitute the evaluation set. Figure~\ref{fig:samples_yorkshire} shows representative examples.

\subsection{Baselines}\label{sec:baselines}

We compare ProtoX-AD against representative baselines from three methodological families: (i) shallow anomaly detection methods, 
(ii) a black-box self-supervised deep learning method, and
(iii) an explainable self-supervised method.

\paragraph{Shallow methods}
We use three popular anomaly detection methods: Isolation Forest (IF)~\cite{liu2008isolation}, One-Class SVM (OCSVM)~\cite{scholkopf1999support}, and Local Outlier Factor (LOF)~\cite{breunig2000lof}. They operate directly on the input data without learned feature extraction, serving as non-deep baselines.

\paragraph{Black-box self-supervised method}
We use ProtoX-AD but without the explainability mechanism. The model is trained to jointly perform self-supervised classification using a linear classifier operating on the learned representations, and to reconstruct the original samples from their augmented views to preserve semantic information. 
This baseline serves as the non-explainable counterpart of ProtoX-AD and provides an upper bound on anomaly detection performance.

\paragraph{Explainable self-supervised method}
KMEx~\cite{gautam2023prototypical} serves as an explainable self-supervised baseline, a prototype-based method that builds upon a self-supervised classification black-box model. KMEx induces prototypes by K-means clustering in the latent space of a pre-trained encoder, and explains predictions by associating each input with its nearest prototype. We apply KMEx on top of the self-supervised classification-based black-box model described above, making it directly comparable to ProtoX-AD in terms of both anomaly detection performance and explainability.

Anomaly scores in KMEx are computed as the cross-entropy of the classification prediction for the identity view, following ProtoX-AD. Explanations are obtained by the nearest prototype in the latent space and visualizing the corresponding representative sample in the input space, selected from the augmented views of the normal training samples.

\noindent Further implementation details of the proposed baselines and ProtoX-AD are presented in~\ref{app:details}.

\subsection{Transformation design for self-supervised anomaly detection}

Transformation design is a key component of self-supervised anomaly detection, as it determines the semantic meaning of the augmented views and the resulting latent representations. We consider two complementary approaches.

\paragraph{Manually defined transformations} 
We design dataset-specific transformation sets that reflect the underlying anomaly structure, leveraging domain knowledge and prior work when available. This ensures alignment between transformation-induced classes and target anomalies. Specifically, in UMD transformations generate bell-shaped patterns, in GTA they simulate warmer and colder temperature anomalies, and in the Yorkshire dataset they mimic leak-related patterns. Full details of all transformations for each dataset are provided in~\ref{app:transformations}.

\paragraph{Learnable neural transformations}
We consider learnable neural transformations modeled as convolutional neural networks, following the approach proposed in NeuTraL AD~\cite{qiu2021neural}. These transformations preserve the dimensionality of the input data, allowing augmented views to be visualized in the original input space. They are learned in an anomaly-agnostic manner, without incorporating prior knowledge about anomaly structure.

\subsection{Evaluation protocol}

The area under the receiver operating characteristic curve (AUROC) and the area under the precision--recall curve (AUPR)~\cite{sanchez2025neucoreclass} are used for anomaly detection performance evaluation. AUROC assesses the ability to discriminate between normal and anomalous samples across decision thresholds, while AUPR is particularly informative in highly imbalanced settings, as it measures the trade-off between precision and recall. Higher values are better.

Explanation quality is assessed both qualitatively, through visual inspection of the explanations, and quantitatively by measuring the similarity between each test sample and its assigned explanation in the input space using Mean Absolute Error (MAE) and Mean Squared Error (MSE).

All experiments are repeated five times using different random seeds, and results are reported as mean and standard deviation.

\section{Experimental Results} \label{sec:experiments}

This section presents a comprehensive evaluation of ProtoX-AD in terms of anomaly detection performance and explainability. For ProtoX-AD, we consider both manually designed transformations and learnable neural transformations to analyze the impact of transformation design on both detection performance and explanation quality.

\subsection{Evaluation of anomaly detection performance}

Tables~\ref{tab:umd_results}, \ref{tab:temperature_results}, and \ref{tab:yorkshire_results} summarize the anomaly detection performance in terms of AUROC and AUPR for the UMD, GTA, and Yorkshire datasets, respectively. Results are averaged over random seeds for UMD, reported separately for the GISTEMP and gcag sources for GTA, and aggregated over seeds, days, and DMAs across four percentile thresholds for Yorkshire.

\paragraph{Manual Transformations}
ProtoX-AD with manually designed transformations achieves strong overall performance across datasets and evaluation metrics. It consistently outperforms shallow methods, including Isolation Forest, LOF, and One-Class SVM, in most cases. Compared to the explainable self-supervised baseline KMEx, which also relies on manually designed transformations, ProtoX-AD attains comparable performance on UMD, slightly improves results on the GISTEMP source of the GTA dataset and across all settings of the Yorkshire dataset, while showing lower performance on the gcag source of GTA. Furthermore, it matches or slightly improves the performance of black-box self-supervised methods, remaining competitive across all percentile thresholds of the Yorkshire dataset.

\paragraph{Neural Transformations}
The use of learnable neural transformations leads to a consistent drop in performance across datasets and evaluation metrics compared to ProtoX-AD with manually designed transformations. While this variant remains competitive with shallow methods on the UMD dataset, it generally underperforms them, particularly on the GTA and Yorkshire datasets. This degradation is consistent across all datasets and percentile thresholds, and is accompanied by increased variability across runs.

\begin{table*}[t]
\centering
\caption{UMD results (average $\pm$ std over seeds).}
\label{tab:umd_results}
\resizebox{\linewidth}{!}{%
\begin{tabular}{lccccccc}
\toprule
 & \multicolumn{3}{c}{\textbf{Shallow methods}} 
 & \multicolumn{3}{c}{\textbf{Explainable SS}} 
 & \textbf{Black-box SS} \\
\cmidrule(lr){2-4} \cmidrule(lr){5-7} \cmidrule(lr){8-8}
\textbf{Metric} 
& IsoForest & LocalOF & OneClassSVM 
& KMEx (MT)& ProtoX-AD (MT) & ProtoX-AD (NT)
& Black-box (MT)\\
\midrule
AUROC (\%) 
& 76.54 $\pm$ 0.71 
& 77.58 $\pm$ 0.00 
& 91.75 $\pm$ 0.00 
& \textbf{100.00 $\pm$ 0.00} 
& \textbf{100.00 $\pm$ 0.00} 
& 83.04 $\pm$ 4.82
& 100.00 $\pm$ 0.00 \\
AUPR (\%) 
& 88.10 $\pm$ 1.29 
& 90.12 $\pm$ 0.00 
& 95.71 $\pm$ 0.00 
& \textbf{100.00 $\pm$ 0.00} 
& \textbf{100.00 $\pm$ 0.00} 
& 92.67 $\pm$ 1.06 
& 100.00 $\pm$ 0.00 \\
\bottomrule
\end{tabular}
}
\begin{minipage}{\columnwidth}
\small  MT: Manual Transformations; NT: Neural Transformations
\end{minipage}
\end{table*}

\begin{table*}[t]
\centering
\caption{Global Temperature Anomalies results for sources GISTEMP and gcag (average $\pm$ std over seeds).}
\label{tab:temperature_results}
\resizebox{\linewidth}{!}{%
\begin{tabular}{lccccccc}
\toprule
 & \multicolumn{3}{c}{\textbf{Shallow methods}} 
 & \multicolumn{3}{c}{\textbf{Explainable SS}} 
 & \textbf{Black-box SS} \\
\cmidrule(lr){2-4} \cmidrule(lr){5-7} \cmidrule(lr){8-8}
\textbf{Metric} 
& IsoForest & LocalOF & OneClassSVM 
& KMEx (MT) & ProtoX-AD (MT) & ProtoX-AD (NT)
& Black-box (MT) \\
\midrule
\multicolumn{7}{l}{\textit{Source GISTEMP}} \\
\midrule
AUROC (\%) & $89.22 \pm 4.87$ & $90.67 \pm 0.00$ & $91.23 \pm 0.00$ & $99.53 \pm 0.37$ & $\mathbf{99.61 \pm 0.44}$ & $83.89 \pm 8.86$ &$99.51 \pm 0.45$ \\
AUPR (\%) & $97.28 \pm 1.32$ & $97.75 \pm 0.00$ & $97.83 \pm 0.00$ & $99.89 \pm 0.09$ & $\mathbf{99.90 \pm 0.11}$ & $96.27 \pm 2.23$ & $99.88 \pm 0.11$ \\
\midrule
\multicolumn{7}{l}{\textit{Source gcag}} \\
\midrule
AUROC (\%) & $79.06 \pm 5.86$ & $82.32 \pm 0.00$ & $86.69 \pm 0.00$ & $\mathbf{96.46 \pm 2.94}$ & $91.47 \pm 0.63$ & $56.57 \pm 26.98 $& $97.49 \pm 1.27$ \\
AUPR (\%) & $96.60 \pm 1.10$ & $97.36 \pm 0.00$ & $98.06 \pm 0.00$ & $\mathbf{99.50 \pm 0.45}$ & $98.67 \pm 0.10$ & $92.18 \pm 5.26 $ & $99.65 \pm 0.20$ \\
\bottomrule
\end{tabular}
}
\begin{minipage}{\columnwidth}
\small  MT: Manual Transformations; NT: Neural Transformations
\end{minipage}
\end{table*}

\begin{table*}[t]
\centering
\caption{Yorkshire Water Leak Detection results for different percentile values (mean $\pm$ std over seeds, days, and DMAs).}
\label{tab:yorkshire_results}
\resizebox{\linewidth}{!}{%
\begin{tabular}{lccccccc}
\toprule
 & \multicolumn{3}{c}{\textbf{Shallow methods}} 
 & \multicolumn{3}{c}{\textbf{Explainable SS}} 
 & \textbf{Black-box SS} \\
\cmidrule(lr){2-4} \cmidrule(lr){5-7} \cmidrule(lr){8-8}
\textbf{Metric} 
& IsoForest & LocalOF & OneClassSVM 
& KMEx (MT) & ProtoX-AD (MT) & ProtoX-AD (NT)
& Black-box (MT) \\
\midrule
\multicolumn{7}{l}{\textit{p = 0.8}} \\
\midrule
AUROC (\%) & $97.64 \pm 3.22$ & $97.10 \pm 4.11$ & $95.64 \pm 4.61$ & $95.59 \pm 3.92$ & $\mathbf{98.00 \pm 2.89}$ & $46.95 \pm 4.38 $ & $97.15 \pm 3.59$ \\
AUPR (\%) & $98.55 \pm 1.95$ & $98.19 \pm 2.60$ & $97.27 \pm 3.00$ & $96.96 \pm 3.13$ & $\mathbf{98.67 \pm 2.00}$ & $76.60 \pm 5.80$ & $98.30 \pm 1.97$ \\
\midrule
\multicolumn{7}{l}{\textit{p = 0.85}} \\
\midrule
AUROC (\%) & $98.05 \pm 2.58$ & $97.38 \pm 3.30$ & $96.66 \pm 3.89$ & $96.15 \pm 3.25$ & $\mathbf{97.67 \pm 2.85}$ & $48.36 \pm 2.48 $& $96.78 \pm 2.98$ \\
AUPR (\%) & $98.20 \pm 2.52$ & $97.28 \pm 4.11$ & $96.61 \pm 4.68$ & $95.98 \pm 3.91$ & $\mathbf{97.49 \pm 3.88}$ & $73.57 \pm 3.60$ & $96.82 \pm 3.58$ \\
\midrule
\multicolumn{7}{l}{\textit{p = 0.9}} \\
\midrule
AUROC (\%) & $98.78 \pm 2.20$ & $98.78 \pm 2.43$ & $97.93 \pm 3.26$ & $96.90 \pm 3.15$ & $\mathbf{98.81 \pm 2.28}$ & $47.92 \pm 2.69 $ & $97.50 \pm 2.66$ \\
AUPR (\%) & $98.36 \pm 3.00$ & $98.20 \pm 4.07$ & $97.24 \pm 4.68$ & $95.92 \pm 4.26$ & $\mathbf{98.32 \pm 3.84}$ & $ 68.00 \pm 3.44 $& $96.59 \pm 4.02$ \\
\midrule
\multicolumn{7}{l}{\textit{p = 0.95}} \\
\midrule
AUROC (\%) & $99.41 \pm 2.50$ & $99.26 \pm 2.81$ & $99.15 \pm 3.12$ & $97.09 \pm 5.43$ & $\mathbf{98.94 \pm 3.46}$ & $47.23 \pm 3.99 $ & $97.68 \pm 4.76$ \\
AUPR (\%) & $98.60 \pm 5.95$ & $98.06 \pm 7.09$ & $97.35 \pm 10.00$ & $93.47 \pm 12.11$ & $\mathbf{96.80 \pm 10.42}$ & $59.74 \pm 4.48$ & $94.23 \pm 12.49$ \\
\bottomrule
\end{tabular}
}
\begin{minipage}{\columnwidth}
\small  MT: Manual Transformations; NT: Neural Transformations
\end{minipage}
\end{table*}

\subsection{Evaluation of explainability}\label{sec:results_explainability}

We evaluate the quality of the explanations provided by ProtoX-AD compared with KMEx from both qualitative and quantitative perspectives, focusing on prototype-based explanations.

\subsubsection{Qualitative evaluation}

\begin{figure}[t]
    \centering
    
    \includegraphics[width=\linewidth]{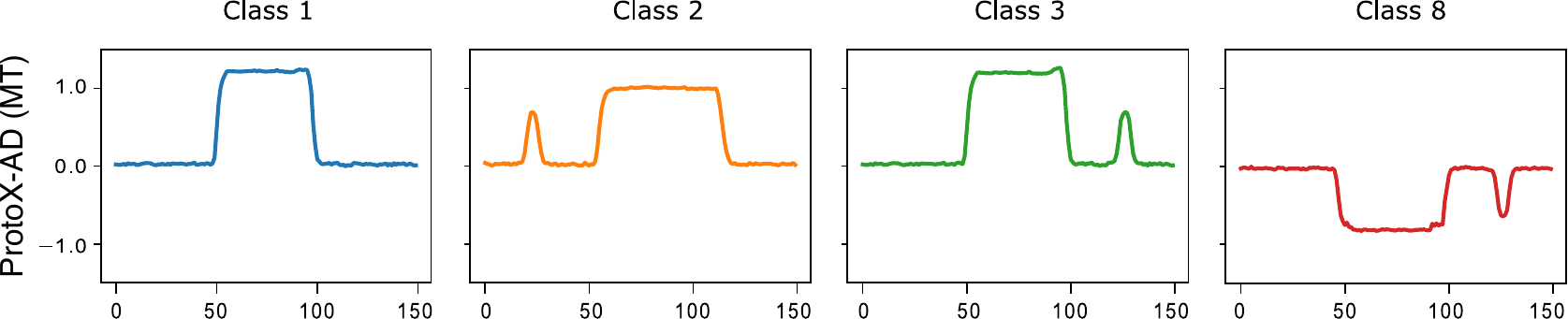}
    
    
    {\color{gray!50}\rule{\linewidth}{0.4pt}}
    
    \vspace{0.9em}
    
    \includegraphics[width=\linewidth]{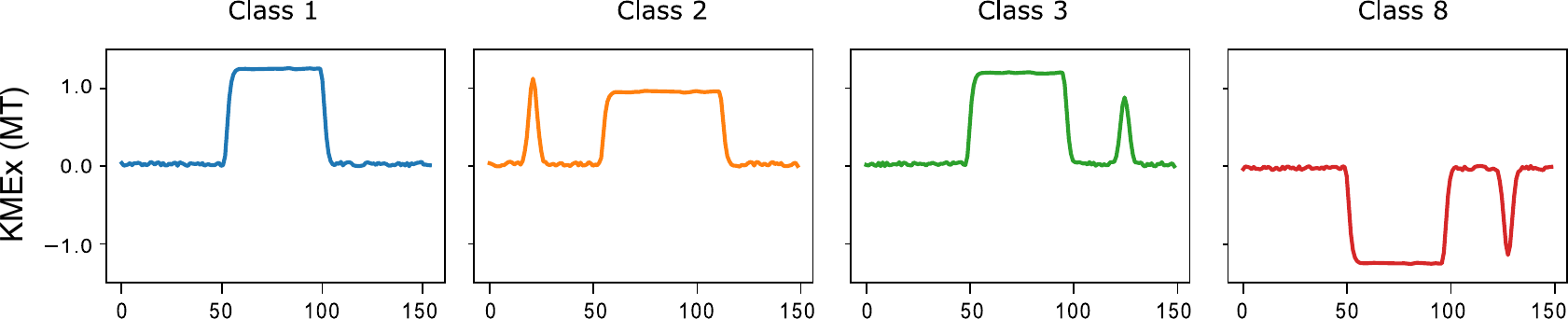}

    {\color{gray!50}\rule{\linewidth}{0.4pt}}
    
    \vspace{0.9em}
    
    \includegraphics[width=\linewidth]{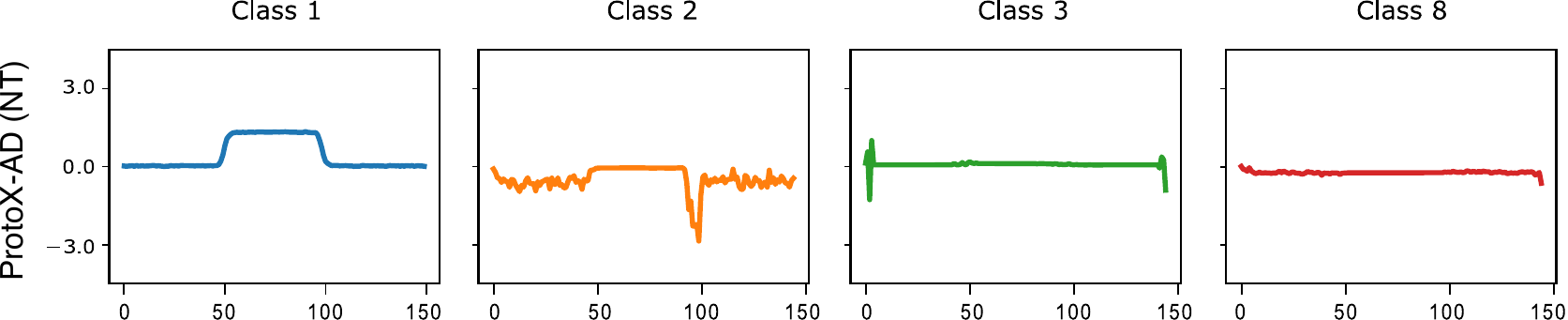}  
    \caption{Learned prototypes for the UMD dataset. Columns represent transformation-induced classes. Rows correspond to ProtoX-AD with manually designed transformations (MT), KMEx (also based on manually defined transformations), and ProtoX-AD with neural transformations (NT), respectively. Colors denote different classes, with blue indicating the identity (normal) class.}
    \label{fig:umd_prototypes}
\end{figure}

\begin{figure}[t]
    \centering
    
    \includegraphics[width=\linewidth]{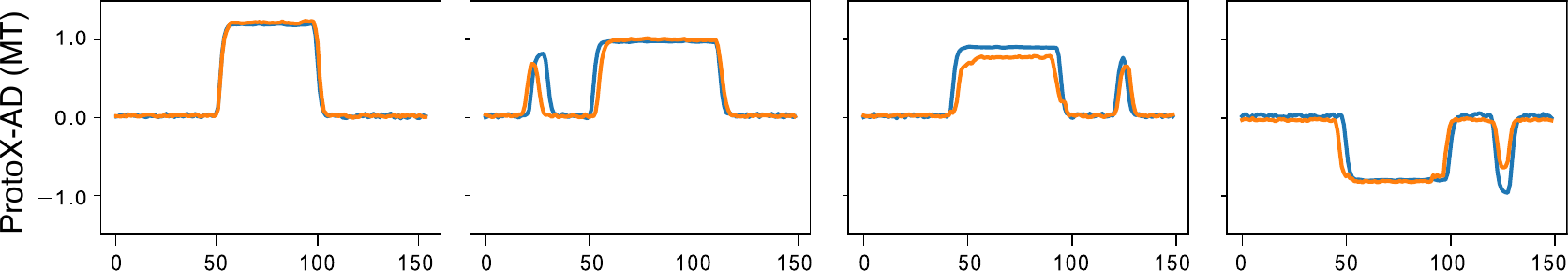}
    
    
    {\color{gray!50}\rule{\linewidth}{0.4pt}}
    
    \vspace{0.9em}
    
    \includegraphics[width=\linewidth]{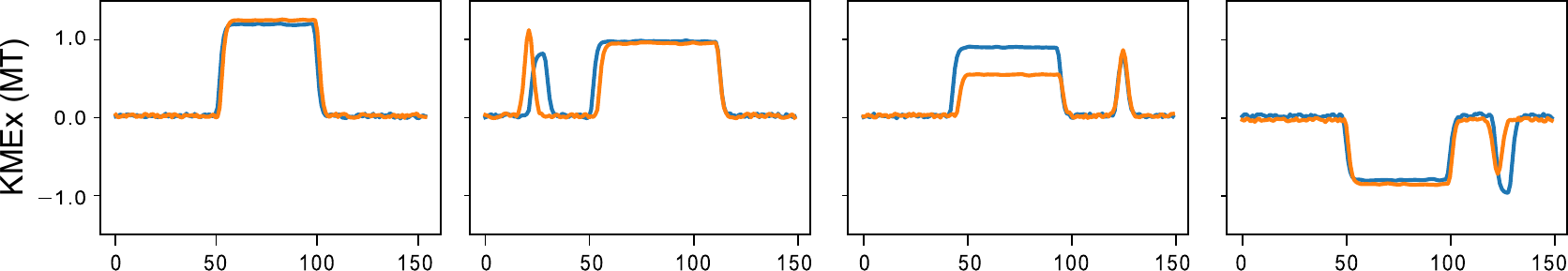}

        {\color{gray!50}\rule{\linewidth}{0.4pt}}
    
    \vspace{0.9em}
    
    \includegraphics[width=\linewidth]{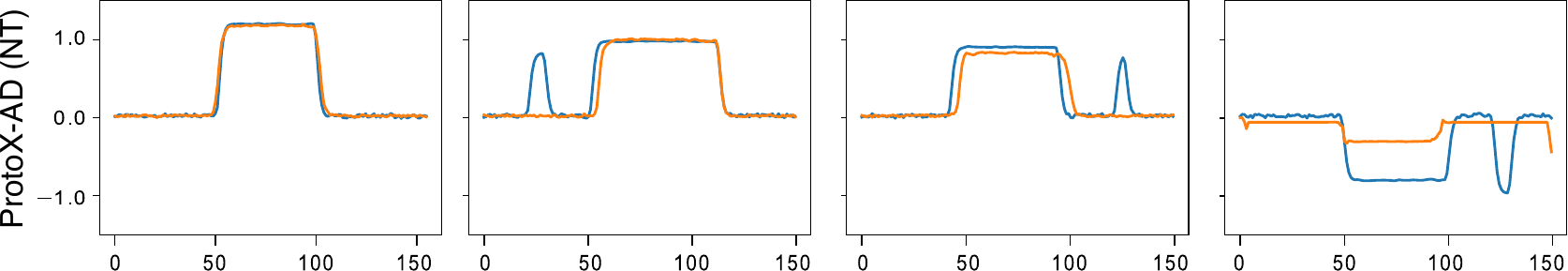}
    
    \caption{Prototype-based explanations for representative test samples from the UMD dataset. Columns correspond to transformation-induced classes. Rows represent ProtoX-AD with manually designed transformations (MT), KMEx (manual transformations), and ProtoX-AD with neural transformations (NT), respectively. Blue curves denote test samples, while orange curves denote the corresponding prototype-based explanations.}
    
    \label{fig:umd_explanations}
\end{figure}

We analyze the learned prototypes and the resulting explanations for representative samples. Figure~\ref{fig:umd_prototypes} illustrates the prototypes learned by each method for the transformation-induced classes considered in the UMD dataset. In addition, Figure~\ref{fig:umd_explanations} presents representative explanations, showing test samples alongside the corresponding explanations assigned by each method. For visualization purposes, we consider a representative subset of four out of the nine classes in this dataset and display one prototype per transformation-induced class.

\paragraph{Manual Transformations} The learned prototypes provide meaningful representations of the transformation-induced classes for both ProtoX-AD and KMEx in the UMD dataset. Each prototype corresponds to a transformation-induced pattern defined by the transformation module: normal plateau behavior (identity), bell-shaped upward deviations at the beginning and end of the sequence without inversion, and bell-shaped downward deviations with inversion of the central plateau. These prototypes therefore capture not only the presence of a bell-shaped perturbation, but also its position and orientation, which define distinct anomalous profiles in this dataset. Generally, ProtoX-AD prototypes appear smoother and reflect the overall structure of each class, whereas KMEx prototypes are sharper and exhibit more localized variations (e.g., more abrupt and pronounced peaks in the bell-shaped patterns).

Both ProtoX-AD and KMEx provide informative prototype-based explanations for representative normal and anomalous samples. The first sample is associated with an identity prototype, correctly reflecting normal plateau behavior. The remaining samples, which exhibit bell-shaped deviations, are matched with prototypes corresponding to the relevant transformation-induced classes, including upward deviations at the beginning or end of the sequence and downward deviations with inversion of the central plateau. These examples show that the explanations are consistent with the learned prototypes and capture the underlying transformation-induced concepts.

\paragraph{Neural Transformations} While the prototype associated with the normal plateau exhibits a clear and stable structure, the remaining prototypes do not resemble meaningful anomaly patterns. Instead of capturing coherent bell-shaped variations or structured deviations of the signal, they exhibit noisy and unstructured variations with no clear correspondence to the transformation-induced classes.

Regarding the explanations, the lack of structure in the learned prototypes is reflected in the resulting explanations. The nearest prototypes assigned to anomalous samples do not consistently correspond to meaningful transformation-induced concepts. In particular, various anomalous samples are associated with the prototype of the identity (normal) class, indicating that the model fails to distinguish between normal and anomalous patterns in terms of the learned prototypes.

Additional qualitative examples for the GISTEMP problem of the GTA dataset are provided in~\ref{app:additional} to further analyze the intra-class diversity of the learned prototypes.

\subsubsection{Quantitative evaluation}

To complement the qualitative analysis, Tables~\ref{tab:explain_umd}, \ref{tab:explain_temperature}, and \ref{tab:explain_yorkshire} report MAE and MSE between each sample and its corresponding explanation for ProtoX-AD (with both manually defined and neural transformations) and KMEx across the UMD, GTA, and Yorkshire datasets. The evaluation considers the use of different numbers of prototypes, namely 1, 3, 5, and 7.

\begin{table*}[t]
\centering
\scriptsize
\caption{Explanation errors ($\times100$) in UMD with 1, 3, 5 and 7 prototypes (mean $\pm$ std over seeds).}
\label{tab:explain_umd}

{
\setlength{\tabcolsep}{4pt}

\resizebox{\linewidth}{!}{%
\begin{tabular}{lcccccccc}
\toprule
 & \multicolumn{2}{c}{\textbf{1P}} & \multicolumn{2}{c}{\textbf{3P}} &
   \multicolumn{2}{c}{\textbf{5P}} & \multicolumn{2}{c}{\textbf{7P}} \\
\cmidrule(lr){2-3} \cmidrule(lr){4-5} \cmidrule(lr){6-7} \cmidrule(lr){8-9}
\textbf{Method} & \textbf{MAE} & \textbf{MSE} & \textbf{MAE} & \textbf{MSE} & \textbf{MAE} & \textbf{MSE} & \textbf{MAE} & \textbf{MSE} \\
\midrule
KMEx (MT)
& $14.76 \pm 0.19$ & $26.06 \pm 0.30$
& $12.84 \pm 0.41$ & $23.83 \pm 0.60$
& $11.92 \pm 0.44$ & $22.26 \pm 0.42$
& $11.01 \pm 0.23$ & $20.56 \pm 0.38$ \\

ProtoX-AD (MT)
& $\boldsymbol{13.89 \pm 0.06}$ & $\boldsymbol{24.62 \pm 0.07}$
& $\boldsymbol{12.32 \pm 0.41}$ & $\boldsymbol{21.62 \pm 0.69}$
& $\boldsymbol{11.14 \pm 0.30}$ & $\boldsymbol{20.20 \pm 0.17}$
& $\boldsymbol{10.99 \pm 0.20}$ & $\boldsymbol{19.75 \pm 0.62}$ \\

ProtoX-AD (NT) & $20.83 \pm 3.18$ & $ 32.60 \pm 2.56 $ & $19.93 \pm 4.07$ & $31.13 \pm 4.29$ & $23.17 \pm 9.19$ & $36.95 \pm 10.47$ & $20.30 \pm 4.09$ & $32.25 \pm 4.93$\\
\bottomrule
\end{tabular}%
}

} 
\vspace{1pt}
{\footnotesize MT: Manual Transformations; NT: Neural Transformations.}
\end{table*}

\begin{table*}[t]
\centering
\scriptsize
\caption{Explanation errors ($\times100$) in Global Temperature Anomalies with 1, 3, 5 and 7 prototypes (mean $\pm$ std over seeds).}
\label{tab:explain_temperature}

\setlength{\tabcolsep}{4pt}
\resizebox{\linewidth}{!}{%
\begin{tabular}{llcccccccc}
\toprule
\multirow{2}{*}{\textbf{Source}} & \multirow{2}{*}{\textbf{Method}}
& \multicolumn{2}{c}{\textbf{1P}}
& \multicolumn{2}{c}{\textbf{3P}}
& \multicolumn{2}{c}{\textbf{5P}}
& \multicolumn{2}{c}{\textbf{7P}} \\
\cmidrule(lr){3-4} \cmidrule(lr){5-6} \cmidrule(lr){7-8} \cmidrule(lr){9-10}
& & \textbf{MAE} & \textbf{MSE} & \textbf{MAE} & \textbf{MSE} & \textbf{MAE} & \textbf{MSE} & \textbf{MAE} & \textbf{MSE} \\
\midrule

\multirow{3}{*}{\textbf{GISTEMP}}
& KMEx (MT)
& $18.58 \pm 1.69$ & $21.66 \pm 1.61$
& $12.97 \pm 1.36$ & $15.96 \pm 1.41$
& $13.07 \pm 1.08$ & $15.80 \pm 0.94$
& $11.08 \pm 0.33$ & $13.77 \pm 0.42$ \\

& ProtoX-AD (MT)
& $\boldsymbol{16.71 \pm 0.13}$ & $\boldsymbol{18.98 \pm 0.15}$
& $\boldsymbol{10.19 \pm 0.11}$ & $\boldsymbol{12.60 \pm 0.13}$
& $\boldsymbol{9.65 \pm 0.05}$  & $\boldsymbol{12.01 \pm 0.08}$
& $\boldsymbol{9.40 \pm 0.12}$  & $\boldsymbol{11.76 \pm 0.12}$ \\

& ProtoX-AD (NT)
& $17.73 \pm 2.09$ & $20.08 \pm 2.01$
& $16.09 \pm 7.19$ & $18.78 \pm 7.55$
& $15.15 \pm 6.00$ & $17.79 \pm 6.41$
& $15.06 \pm 7.46$ & $17.81 \pm 8.17$ \\
\midrule

\multirow{3}{*}{\textbf{gcag}}
& KMEx (MT)
& $17.29 \pm 0.73$ & $20.16 \pm 0.89$
& $14.04 \pm 1.26$ & $16.97 \pm 1.04$
& $13.11 \pm 1.42$ & $15.68 \pm 1.41$
& $13.15 \pm 2.05$ & $15.89 \pm 1.95$ \\

& ProtoX-AD (MT)
& $\boldsymbol{15.89 \pm 0.10}$ & $\boldsymbol{18.26 \pm 0.10}$
& $\boldsymbol{10.57 \pm 0.08}$ & $\boldsymbol{12.97 \pm 0.05}$
& $\boldsymbol{10.09 \pm 0.05}$ & $\boldsymbol{12.44 \pm 0.05}$
& $\boldsymbol{9.94 \pm 0.10}$  & $\boldsymbol{12.29 \pm 0.11}$ \\

& ProtoX-AD (NT)
& $19.45 \pm 4.82$ & $22.35 \pm 5.49$
& $16.89 \pm 6.69$ & $19.48 \pm 6.92$
& $15.06 \pm 6.02$ & $17.84 \pm 6.67$
& $15.08 \pm 5.84$ & $17.88 \pm 6.27$ \\
\bottomrule
\end{tabular}
}
\vspace{2pt}
{\footnotesize MT: Manual Transformations; NT: Neural Transformations.}
\end{table*}

\begin{table*}[t]
\centering
\scriptsize
\caption{Explanation errors ($\times100$) in Yorkshire Water Leak Detection with 1, 3, 5 and 7 prototypes (mean $\pm$ std over seeds, days and DMAs).}
\label{tab:explain_yorkshire}

\setlength{\tabcolsep}{4pt}

\resizebox{\linewidth}{!}{%
\begin{tabular}{llcccccccc}
\toprule
\multirow{2}{*}{\textbf{Setting}} & \multirow{2}{*}{\textbf{Method}}
& \multicolumn{2}{c}{\textbf{1P}}
& \multicolumn{2}{c}{\textbf{3P}}
& \multicolumn{2}{c}{\textbf{5P}}
& \multicolumn{2}{c}{\textbf{7P}} \\
\cmidrule(lr){3-4} \cmidrule(lr){5-6} \cmidrule(lr){7-8} \cmidrule(lr){9-10}
& & \textbf{MAE} & \textbf{MSE} & \textbf{MAE} & \textbf{MSE} & \textbf{MAE} & \textbf{MSE} & \textbf{MAE} & \textbf{MSE} \\
\midrule

\multirow{3}{*}{$p = 0.80$}
& KMEx (MT)
& $59.30 \pm 23.44$  & $64.47 \pm 25.21$
& $42.93 \pm 23.88$ & $48.15 \pm 25.34$
& $38.79 \pm 23.19$  & $44.09 \pm 24.39$
& $37.90 \pm 22.20$ & $43.11 \pm 23.34$ \\

& ProtoX-AD (MT)
& $\boldsymbol{47.22 \pm 22.16}$ & $\boldsymbol{51.07 \pm 23.41}$
& $\boldsymbol{34.08 \pm 20.24}$ & $\boldsymbol{38.00 \pm 20.76}$
& $\boldsymbol{33.26 \pm 19.69}$ & $\boldsymbol{37.17 \pm 20.24}$
& $\boldsymbol{32.62 \pm 19.17}$ & $\boldsymbol{36.57 \pm 19.80}$ \\

& ProtoX-AD (NT)
& $77.76 \pm 40.07$  & $80.78 \pm 40.79$
& $69.89 \pm 37.61$ & $73.06 \pm 38.14$
& $67.59 \pm 37.18$ & $70.79 \pm 37.63$
& $66.51 \pm 36.67$ & $69.74 \pm 37.09$ \\
\midrule

\multirow{3}{*}{$p = 0.85$}
& KMEx (MT)
& $61.76 \pm 26.40$ & $66.86 \pm 27.86$
& $44.12 \pm 25.54$ & $49.43 \pm 26.96$
& $40.79 \pm 24.36$ & $46.12 \pm 25.60$
& $39.31 \pm 24.61$ & $44.52 \pm 25.69$ \\

& ProtoX-AD (MT)
& $\boldsymbol{49.22 \pm 24.18}$ & $\boldsymbol{53.07 \pm 25.20}$
& $\boldsymbol{35.35 \pm 23.07}$ & $\boldsymbol{39.26 \pm 23.40}$
& $\boldsymbol{34.21 \pm 22.32}$ & $\boldsymbol{38.05 \pm 22.63}$
& $\boldsymbol{33.53 \pm 21.95}$ & $\boldsymbol{37.39 \pm 22.28}$ \\

& ProtoX-AD (NT)
& $83.68 \pm 44.32$ & $86.65 \pm 44.90$
& $74.03 \pm 41.81$ & $77.15 \pm 42.13$
& $71.66 \pm 41.39$ & $74.84 \pm 41.69$
& $70.75 \pm 40.83$ & $73.93 \pm 41.08$ \\
\midrule

\multirow{3}{*}{$p = 0.90$}
& KMEx (MT)
& $59.93 \pm 28.06$ & $65.08 \pm 29.29$
& $44.00 \pm 26.28$ & $49.25 \pm 27.43$
& $40.14 \pm 26.24$ & $45.29 \pm 27.31$
& $37.92 \pm 24.29$ & $43.09 \pm 25.41$ \\

& ProtoX-AD (MT)
& $\boldsymbol{48.19 \pm 24.91}$ & $\boldsymbol{51.97 \pm 25.64}$
& $\boldsymbol{35.02 \pm 24.12}$ & $\boldsymbol{38.89 \pm 24.34}$
& $\boldsymbol{33.83 \pm 24.56}$ & $\boldsymbol{37.65 \pm 24.73}$
& $\boldsymbol{33.18 \pm 23.95}$ & $\boldsymbol{36.99 \pm 24.23}$ \\

& ProtoX-AD (NT)
& $83.97 \pm 44.91$  & $86.94 \pm 45.35$
& $73.73 \pm 42.90$ & $76.87 \pm 43.03$
& $70.71 \pm 42.42$ & $73.86 \pm 42.48$
& $69.41 \pm 41.77$ & $72.57 \pm 41.81$ \\
\midrule

\multirow{3}{*}{$p = 0.95$}
& KMEx (MT)
& $57.62 \pm 31.39$ & $62.45 \pm 31.80$
& $44.57 \pm 31.81$ & $50.24 \pm 34.47$
& $39.20 \pm 26.07$ & $44.58 \pm 27.91$
& $42.57 \pm 44.98$ & $42.57 \pm 44.98$ \\

& ProtoX-AD (MT)
& $\boldsymbol{49.14 \pm 30.94}$ & $\boldsymbol{52.75 \pm 31.40}$
& $\boldsymbol{36.14 \pm 30.74}$ & $\boldsymbol{39.87 \pm 30.75}$
& $\boldsymbol{33.96 \pm 28.55}$ & $\boldsymbol{37.67 \pm 28.58}$
& $\boldsymbol{32.50 \pm 26.86}$ & $\boldsymbol{36.19 \pm 26.97}$ \\

& ProtoX-AD (NT)
& $86.37 \pm 49.95$ & $89.38 \pm 50.18$
& $74.93 \pm 48.34$ & $78.08 \pm 48.26$
& $71.61 \pm 47.84$ & $74.74 \pm 47.71$
& $69.95 \pm 47.64$ & $73.10 \pm 47.50$ \\
\bottomrule
\end{tabular}
}
\vspace{2pt}
{\footnotesize MT: Manual Transformations; NT: Neural Transformations.}
\end{table*}

\paragraph{Manual Transformations}
Across all datasets and experimental settings, ProtoX-AD achieves lower explanation errors than KMEx, reflecting a closer match between the learned prototypes and the explained samples in the input space. For both methods, increasing the number of prototypes leads to lower errors, as a richer prototype set enables more fine-grained explanations. For any number of prototypes, ProtoX-AD consistently outperforms KMEx.

\paragraph{Neural Transformations} 
ProtoX-AD with neural transformations exhibits substantially higher explanation errors across all datasets and prototype configurations. This degradation is consistent across all prototype settings, with the neural variant underperforming both ProtoX-AD and KMEx when using manually defined transformations. Although increasing the number of prototypes leads to a slight reduction in error, the gap with manually defined transformations remains large, indicating that a larger prototype set does not compensate for the lack of meaningful structure in the learned transformations.

\section{Discussion}\label{sec:neural}

We discuss the main findings of our experimental evaluation, focusing on three main aspects: the detection performance of explainable models, the comparison between ProtoX-AD and KMEx through their prototype-based explanations, and the impact of transformation design on SSC-TSAD , considering both manually defined and learnable neural transformations.

\subsection{Detection Performance of Explainable Models}

We analyze how incorporating explainability affects anomaly detection performance in the setting with manually defined transformations, comparing explainable methods with their black-box counterpart and evaluating the relative performance of ProtoX-AD and KMEx.

\paragraph{Explainability does not inherently degrade anomaly detection performance}
Introducing explicit explainability mechanisms is often assumed to compromise detection performance when compared to black-box models. However, our results show that this effect is not systematic. On the synthetic UMD dataset (Table~\ref{tab:umd_results}), both self-explainable methods achieve perfect detection performance, indicating no degradation. On the GTA dataset (Table~\ref{tab:temperature_results}), the impact varies across sources: explainable methods remain competitive with the black-box baseline on GISTEMP, while showing a performance drop on gcag, particularly for ProtoX-AD. In contrast, on the Yorkshire dataset (Table~\ref{tab:yorkshire_results}), ProtoX-AD consistently matches or outperforms the black-box baseline across all settings, whereas KMEx exhibits a more noticeable gap. Overall, these results indicate that incorporating explainability does not inherently degrade detection performance, although its impact depends on the dataset and the specific modeling approach.

\paragraph{ProtoX-AD exhibits more consistent detection performance than KMEx}
A direct comparison between the two explainable methods reveals a non-uniform but interpretable pattern. On UMD (Table \ref{tab:umd_results}), both approaches perform identically due to the simplicity of the anomaly structure. On GISTEMP, ProtoX-AD achieves slightly stronger detection performance, whereas on gcag KMEx shows higher AUROC and AUPR values (Table \ref{tab:temperature_results}. However, on the Yorkshire dataset (Table \ref{tab:yorkshire_results}), which comprises multiple percentile thresholds and operational scenarios, ProtoX-AD consistently outperforms KMEx across all settings. Taken together, these results indicate that while neither method universally dominates, ProtoX-AD exhibits more consistent performance across datasets and evaluation settings, suggesting that integrating prototypes directly into the learning objective provides a more stable detection mechanism.

\subsection{Comparison of Explainable Methods through Prototype-Based Explanations}

We compare ProtoX-AD and KMEx in terms of the structure and consistency of their prototype-based explanations when using manually defined transformations.

\paragraph{ProtoX-AD learns smoother and more representative class-level prototypes}

A qualitative difference can be observed between the prototypes produced by ProtoX-AD and KMEx. ProtoX-AD yields smoother prototypes that capture class-level structure, whereas KMEx prototypes tend to reflect instance-specific variations. This difference stems from how prototypes are constructed: ProtoX-AD learns latent concepts optimized during training and reconstructs them in the input space, while KMEx explanations are obtained by matching individual training instances. Consequently, ProtoX-AD captures more general transformation-induced patterns in its prototypes, whereas KMEx exhibits instance-specific noise. This leads to consistently lower reconstruction errors between samples and their explanations (Tables~\ref{tab:explain_umd}, \ref{tab:explain_temperature}, and \ref{tab:explain_yorkshire}) in the case of ProtoX-AD. Note that increasing the number of prototypes further allows ProtoX-AD to model finer intra-class variability through more specialized concepts.

\begin{figure}[t]
    \centering
    \includegraphics[width=0.48\linewidth]{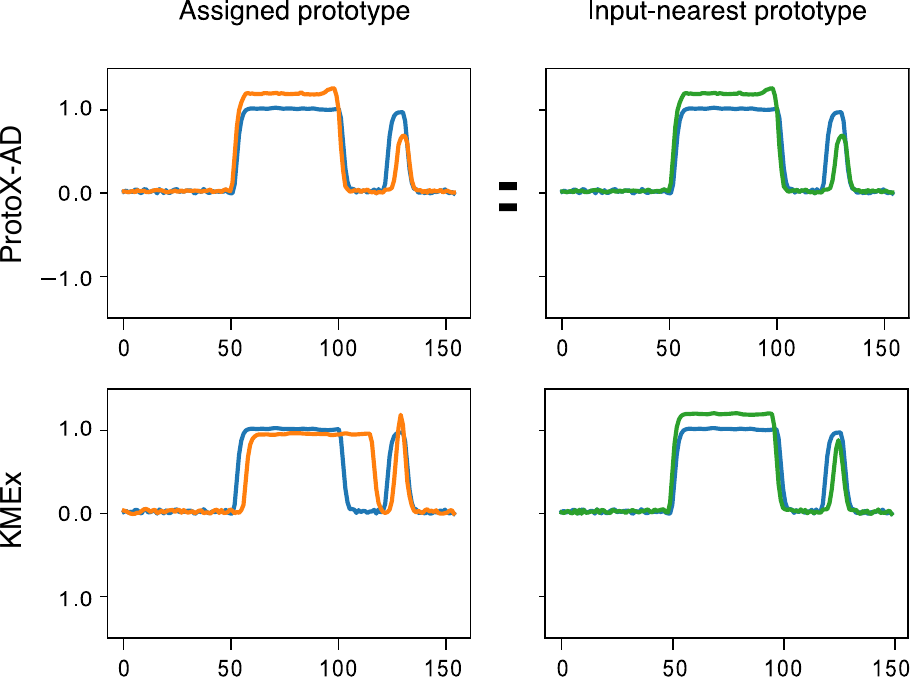}
    \hfill
    \hspace{-0.05pt}
    {\color{gray!50}\vrule}
    \includegraphics[width=0.48\linewidth]{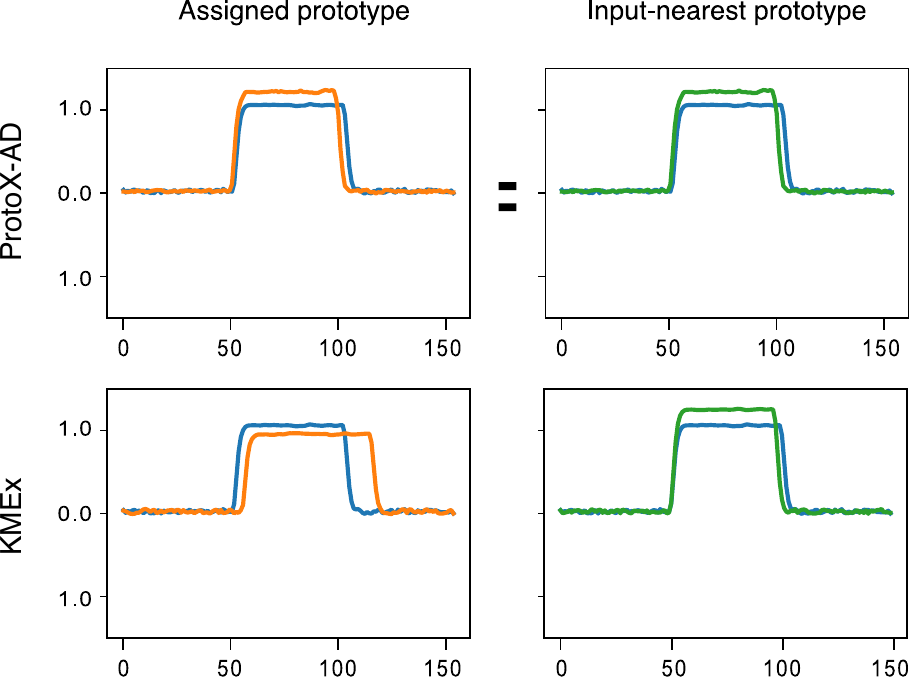}
    \caption{Alignment between latent assignment and input-space similarity in the UMD dataset. For ProtoX-AD (top row), the assigned prototype coincides with the input-space nearest prototype. In contrast, for KMEx (bottom row), the assigned prototype does not necessarily correspond to the nearest reconstructed concept in the input space.}
    \label{fig:alignment_umd}
\end{figure}

\paragraph{ProtoX-AD enforces consistent latent–input alignment}
ProtoX-AD exhibits a consistent alignment between latent-space assignments and their corresponding representations in the input space, whereas such alignment is not guaranteed in KMEx. As illustrated in Figure~\ref{fig:alignment_umd}, the prototype assigned to a sample by ProtoX-AD in the latent space typically coincides with the nearest prototype in the input space. In contrast, KMEx assigns prototypes that do not correspond to the closest input-space concept, leading to discrepancies between latent proximity and semantic similarity.

This behavior is consistent with prior observations~\cite{hoffmann2021looks} that latent similarity does not necessarily imply semantic similarity in prototype-based models. In ProtoX-AD, this alignment is promoted by learning prototype concepts jointly with the model and by the KL-divergence objective, which encourages a coherent organization of the latent space, resulting in more consistent and reliable explanations.

\subsection{On the importance of Transformation Design: Manual vs. Neural Transformations}

We now analyze the role of transformation design in self-supervised anomaly detection by comparing manually defined and learnable neural transformations, focusing on their impact on both detection performance and the quality of the resulting explanations.

\paragraph{The effectiveness of transformation design depends on its alignment with the anomaly structure}
As observed in Section \ref{sec:experiments} (see Tables \ref{tab:umd_results}, \ref{tab:temperature_results} and \ref{tab:yorkshire_results}), manually defined transformations that align with the underlying anomaly structure lead to strong detection performance. In contrast, replacing them with neural transformations consistently degrades performance across datasets, particularly in more complex scenarios where informative augmented views are harder to learn. This highlights the central role of transformation design in self-supervised anomaly detection.

\paragraph{Domain knowledge in transformation design is essential for anomaly characterization}
Beyond detection performance, transformation design also plays a critical role in the interpretability of anomaly explanations. When manually defined transformations are aligned with the underlying anomaly structure, transformation-induced classes correspond to meaningful variations of the signal, enabling the model to associate anomalous samples with interpretable concepts and support anomaly characterization. 

In contrast, neural transformations, which are learned in an anomaly-agnostic manner, fail to capture such semantic structure. As a result, the corresponding prototypes do not align with the target anomalous patterns and fail to specialize in distinct anomaly types. This behavior is illustrated in Figure~\ref{fig:umd_explanations}, where many samples—both normal and anomalous—are matched with prototypes associated with the identity transformation. Consequently, explanations derived from neural transformations do not provide meaningful information for anomaly characterization.

\section{Conclusion}

This work introduces ProtoX-AD, a prototype-based self-supervised framework for explainable time series anomaly detection. By leveraging transformation-induced surrogate classes, the model learns structured representations aligned with interpretable prototypes, enabling anomaly detection together with intrinsic explainability and anomaly characterization.

Experimental results show that incorporating domain knowledge through manually designed transformations is key to achieving strong performance in self-supervised anomaly detection. In this setting, ProtoX-AD achieves detection performance comparable to its black-box counterpart, while outperforming the explainable baseline KMEx and providing more structured and precise explanations. In particular, the learned prototypes capture class-level patterns and yield explanations that are both quantitatively and qualitatively more consistent.

Regarding transformation design, our analysis of manual and neural transformations in SSC-TSAD highlights a trade-off between domain-specific transformation design and learnable augmentation mechanisms. While manual transformations that encode prior knowledge about anomaly structure lead to stronger detection performance and more meaningful explanations, relying on such knowledge may be unrealistic in many real-world scenarios. Conversely, learning neural transformations solely under generic structural constraints is insufficient to produce augmented views that correspond to meaningful anomaly-related variations of the signal, which are necessary both for competitive anomaly detection performance and for the characterization of different anomalous profiles. Future work should therefore explore adaptive approaches that progressively incorporate knowledge about detected anomalies to learn transformations in a more flexible way, enabling both robust detection and meaningful anomaly characterization.

Finally, although this work focuses on time series anomaly detection, the proposed framework is not limited to this setting. By relying on transformation-induced representations and prototype-based learning, ProtoX-AD can be extended to other data modalities. Investigating its applicability across domains constitutes a promising direction for future work.

\subsubsection*{Acknowledgments}
This work was supported by the Research Council of Norway (NFR), through its Centre for Research-based Innovation (grant no. 309439) and FRIPRO (grant nos. 303514 and 360068); and by the Basque Government (grant no. IT1504-22). This publication is part of project PID2022-137442NB-I00 funded by MICIU/AEI/10.13039/501100011033. A. S. F. acknowledges financial support from the Department of Education of the Basque Government (grant no. PRE\_2022\_1\_0103).

\bibliographystyle{elsarticle-num}
\bibliography{bibilography}

\appendix

\section{Manual Transformations per dataset}\label{app:transformations}

Following prior work~\cite{yoo2022data}, we design dataset-specific transformations that generate augmented views that mimic the anomalous behaviors of interest in each problem, together with the identity transformation.

Transformations are stochastically parameterized to introduce variability across augmented views, see Table~\ref{tab:manual_transformation_ranges}.

\begin{table*}[t]
\centering

\caption{Stochastic parameters and sampling ranges used by the manual transformation modules for each dataset. All parameters are resampled at each forward pass of the transformation module.}
\label{tab:manual_transformation_ranges}
\resizebox{0.7\linewidth}{!}{%
\begin{tabular}{llc}
\toprule
\textbf{Transformation} & \textbf{Stochastic parameter} & \textbf{Range} \\
\midrule
\multicolumn{3}{l}{\textit{UMD Dataset}} \\
\midrule
Local bump (all variants) & Amplitude & $[0.75,\;1.25]$ \\
Local bump (all variants) & Inward temporal shift & $[15,\;20]$ \\
Local bump (all variants) & Edge jitter & $[0,\;5]$ \\
\midrule
\multicolumn{3}{l}{\textit{Temperature Dataset}} \\
\midrule
Cold-heavy shift & Target annual mean & $[-1.225,\;-0.80]$ \\
Cold-light shift & Target annual mean & $[-0.70,\;-0.30]$ \\
Warm-light shift & Target annual mean & $[0.30,\;0.70]$ \\
Warm-heavy shift & Target annual mean & $[0.80,\;1.225]$ \\
\midrule
\multicolumn{3}{l}{\textit{Yorkshire Dataset}} \\
\midrule
Low magnitude scaling & Multiplicative factor & $[1.4,\;1.7]$ \\
Medium magnitude scaling & Multiplicative factor & $[2.0,\;2.3]$ \\
High magnitude scaling & Multiplicative factor & $[2.6,\;2.9]$ \\
\bottomrule
\end{tabular}
}
\end{table*}

\paragraph{Transformations for UMD}
For the UMD dataset, transformations are designed to generate augmented views reflecting the two anomalous classes present in the data: upward and downward bell-shaped patterns. In addition to the identity transformation, four transformations per anomaly type are defined by combining the position of the bell (at the beginning or end of the sequence) and the orientation of the central plateau (preserved or inverted), yielding nine transformations in total. The amplitude of the bell, a small edge jitter, and an inward temporal shift are randomly sampled within predefined ranges to introduce variability while preserving the anomaly structure.

\paragraph{Transformations for Global Temperature Anomalies}
For the GTA dataset, transformations consist of global upward or downward shifts applied to normal samples, modifying the overall level of the series to generate colder-than-average and warmer-than-average augmented views consistent with the anomaly definition of the dataset. In addition to the identity transformation, four transformations are defined corresponding to cold-heavy, cold-light, warm-light, and warm-heavy regimes, distinguished by the magnitude of the enforced shift. For each transformation, a fixed target interval for the annual anomaly value is defined, and the applied displacement is randomly sampled from the corresponding interval.

\paragraph{Transformations for Yorkshire Water Leak Detection}

For the Yorkshire Water Leak Detection dataset, transformations follow the approach proposed by~\cite{blazquez2021water}, where anomaly-relevant views are generated by applying multiplicative scaling factors to the original time series. This design is motivated by the physical nature of leak events, which typically manifest as sustained increases in water flow. In addition to the identity transformation, three transformations with increasing scaling magnitudes are defined to simulate leak scenarios of progressively increasing severity. For each transformation, the scaling factor is randomly sampled from a predefined interval corresponding to low, medium, or high magnitude.

\section{Implementation and experimental details}\label{app:details}

Here we describe the implementation choices and training protocols used for ProtoX-AD and all baseline methods.

\paragraph{Implementation of Methods}

All shallow baseline methods are implemented using the scikit-learn library~\cite{pedregosa2011scikit} with default hyperparameters.

We adopt the TS2Vec encoder~\cite{yue2022ts2vec} as backbone architecture for all deep learning methods. For all methods, decoder architectures are symmetric to the encoder. In ProtoX-AD, the encoder outputs the mean and variance parameters required by the variational formulation.

\paragraph{Training details and hyperparameters}
Models are trained for up to 1{,}000 epochs using early stopping, with a \texttt{ReduceLROnPlateau} scheduler that reduces the initial learning rate of $10^{-3}$ when the training loss plateaus (patience of 10 epochs). Training is stopped when the learning rate falls below $10^{-6}$.

We use a batch size of 4 across all datasets. Additionally, for all self-supervised methods, each mini-batch is augmented using 5 stochastic repetitions of the transformation module during training, which can be interpreted as data augmentation in the transformation space and increases intra-class diversity.

Finally, ProtoX-AD and KMEx introduce a hyperparameter controlling the number of prototypes per class, which is fixed to 3.

\section{Analysis of Intra-Class Prototype Diversity}\label{app:additional}

In this appendix, we provide additional qualitative results on the GISTEMP problem of the GTA dataset. While the main paper focuses on representative prototypes and explanations, these figures illustrate how multiple prototypes within the same transformation-induced class capture different variants of the corresponding concept.

\begin{figure}[t]
    \centering
    
    \includegraphics[width=0.675\linewidth]{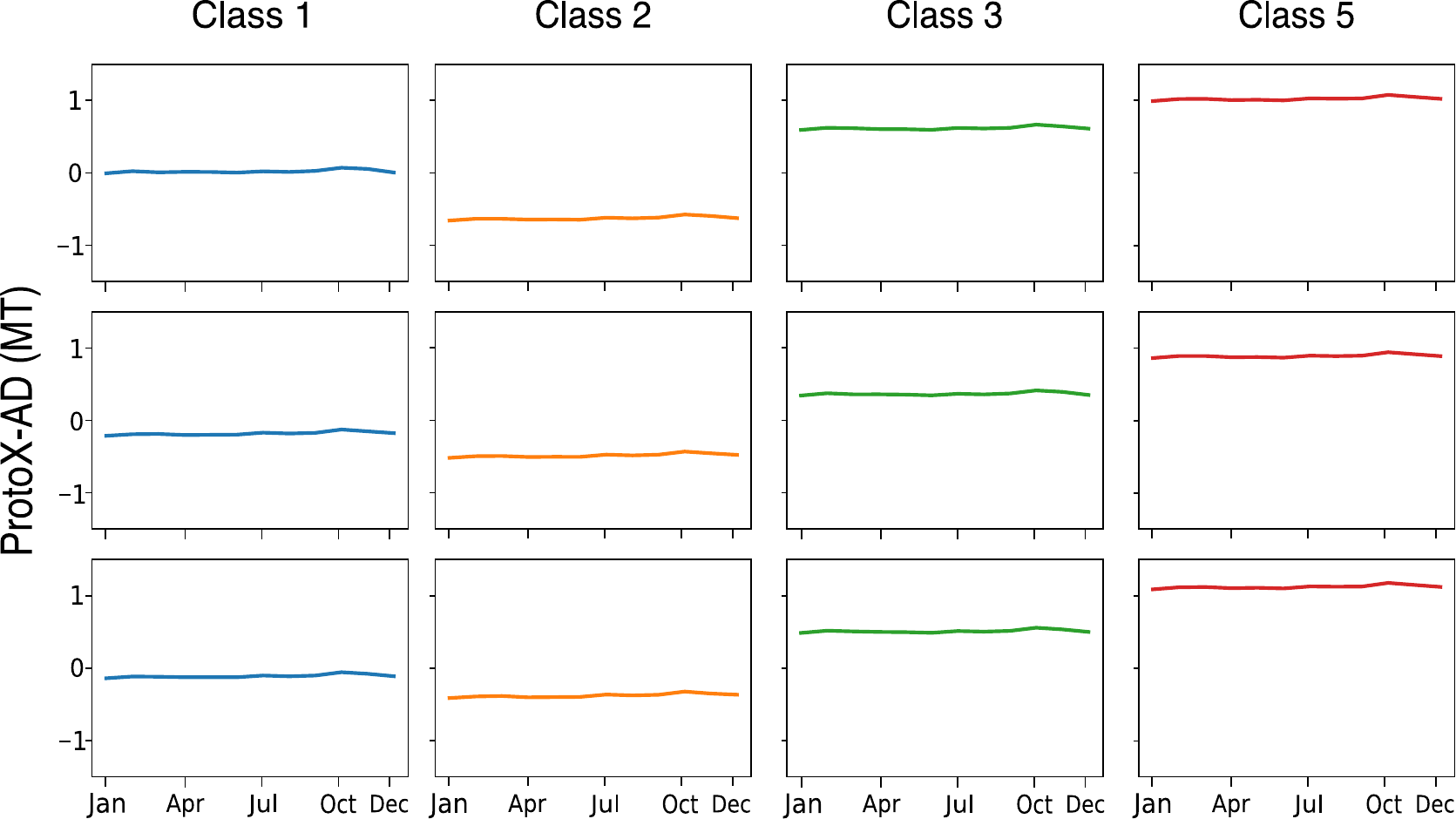}
    
    
    {\color{gray!50}\rule{\linewidth}{0.4pt}}
    
    \vspace{0.5em}
    
    \includegraphics[width=0.675\linewidth]{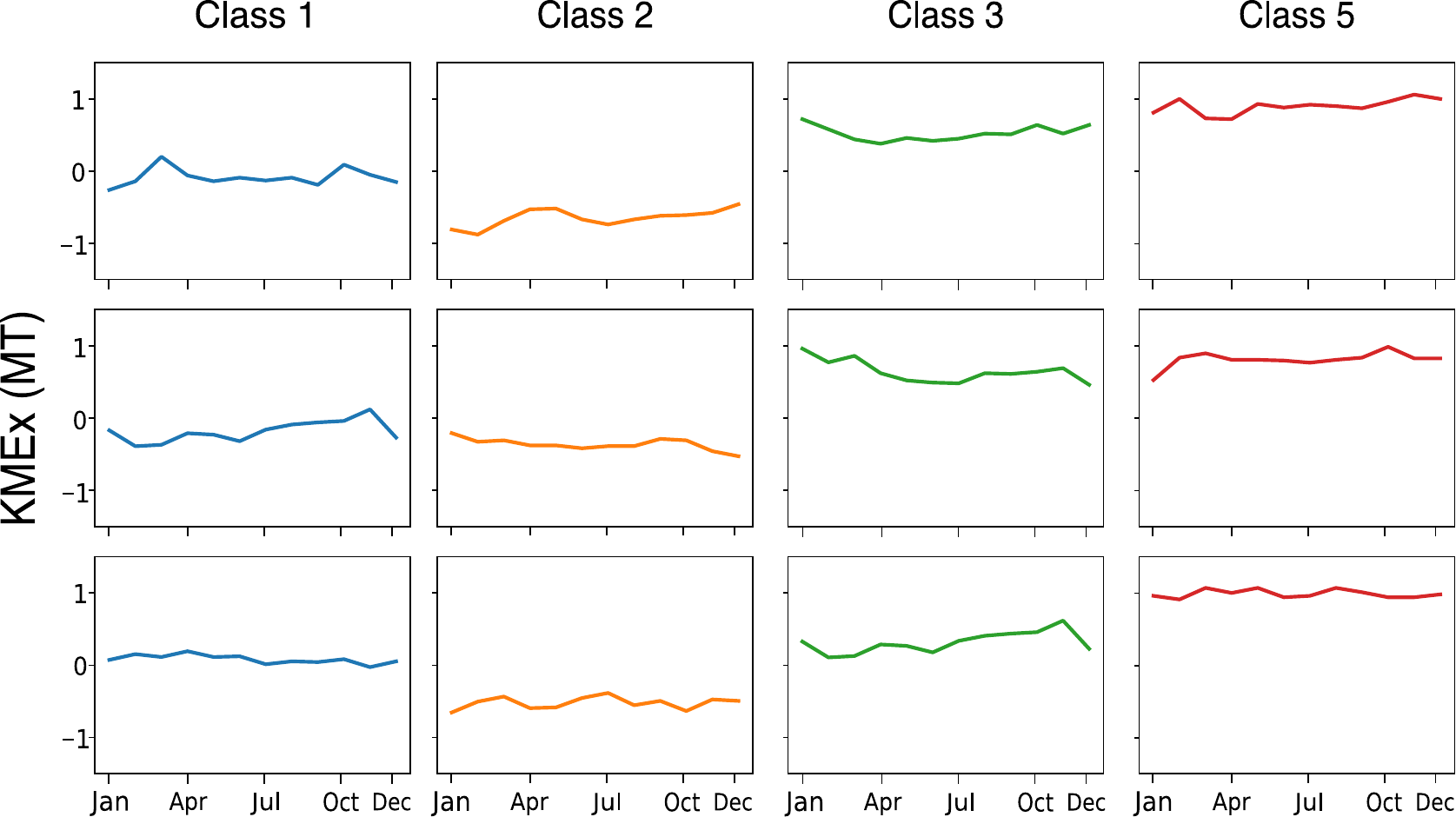}

    {\color{gray!50}\rule{\linewidth}{0.4pt}}
    
    \vspace{0.5em}
    
    \includegraphics[width=0.675\linewidth]{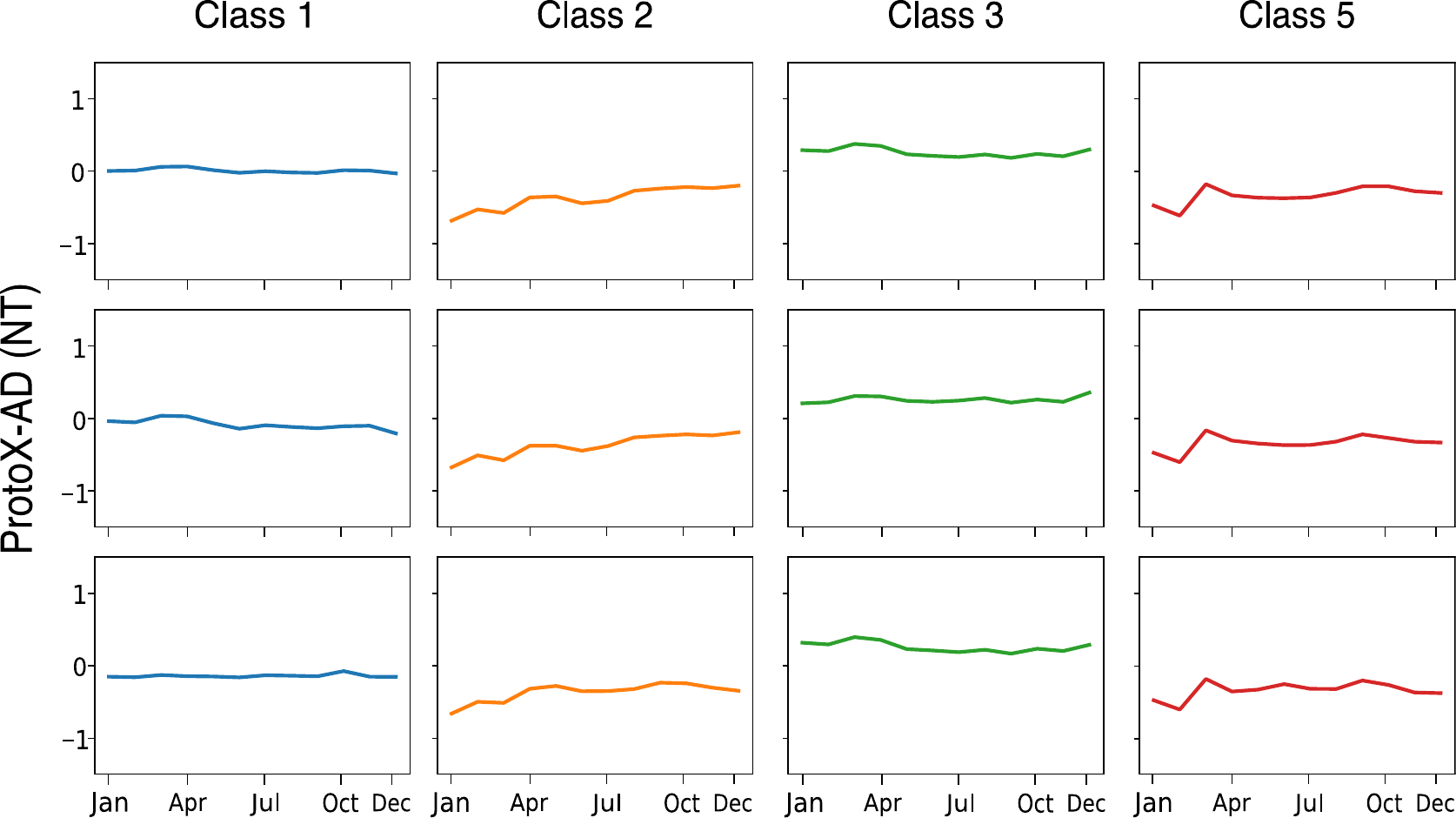}  
    \caption{Learned prototypes for the GISTEMP dataset. Columns represent transformation-induced classes. Rows correspond to ProtoX-AD with manually designed transformations (MT), KMEx (also based on manually defined transformations), and ProtoX-AD with neural transformations (NT), respectively. Colors denote different classes, with blue indicating the identity (normal) class.}
    \label{fig:gistemp_prototypes}
\end{figure}

Figure~\ref{fig:gistemp_prototypes} presents multiple learned prototypes associated with the identity, cold-light, warm-light, and warm-heavy transformation-induced classes, respectively. These visualizations illustrate how the learned prototypes capture different variations within the same transformation-induced concept across methods and transformation designs. 

\paragraph{Manual Transformations}

ProtoX-AD learns multiple prototypes that remain semantically coherent within each transformation-induced class while still capturing meaningful intra-class variability. In particular, the variability among ProtoX-AD prototypes within the same class is mainly reflected through changes in the amplitude of the series, directly corresponding to different degrees of temperature deviation within the same regime. Therefore, the learned prototypes vary primarily along the semantically relevant dimension of the problem while preserving the overall structure of the corresponding transformation-induced concept. In contrast, KMEx prototypes exhibit stronger variability in the shape of the series itself, reflecting a greater dependence on instance-specific variations and noise inherited from individual training samples rather than on coherent class-level concepts.


\paragraph{Neural Transformations}
In contrast to manually designed transformations, the diversity observed with neural transformations is minimal, as the learned augmented views tend to converge to more similar patterns, resulting in nearly redundant prototypes within each class. Consequently, the learned prototypes do not exhibit a clear correspondence with semantically meaningful temperature regimes, and any apparent similarity with specific anomaly patterns is incidental rather than systematically induced by the learned transformations.

This behavior can be explained by the lack of explicit variability mechanisms in the learned transformations. While the reconstruction objective prevents complete latent collapse and the clustering objective encourages representative latent concepts, neural transformations do not incorporate the stochastic transformation parameters used in manually designed transformations (Table~\ref{tab:manual_transformation_ranges}), which naturally generate diverse augmented views. As a result, the variability among learned prototypes remains limited. These observations suggest that introducing explicit variability mechanisms into neural transformations could be an interesting direction for future work.

\end{document}